\documentclass[letterpaper]{article} 
\usepackage{aaai25}  
\usepackage{times}  
\usepackage{helvet}  
\usepackage{courier}  
\usepackage[hyphens]{url}  
\usepackage{graphicx} 
\usepackage{natbib}  
\usepackage{caption} 
\usepackage{algorithm}
\usepackage{algorithmic}

\usepackage{amsmath}
\usepackage{mathtools}
\usepackage{amssymb}
\usepackage{amsthm}
\usepackage{enumitem}
\usepackage{colortbl}

\usepackage{subcaption}

\usepackage{amssymb}
\usepackage{pifont}
\newcommand{\cmark}{\ding{51}}%
\newcommand{\xmark}{\ding{55}}%

\usepackage{multirow} 
\usepackage[capitalize,noabbrev]{cleveref}
\usepackage{graphicx}
\usepackage{multicol}

\usepackage{enumitem}
\usepackage{booktabs}

\usepackage[textsize=tiny]{todonotes}

\usepackage{newfloat}
\usepackage{listings}
\DeclareCaptionStyle{ruled}{labelfont=normalfont,labelsep=colon,strut=off} 
\floatstyle{ruled}
\newfloat{listing}{tb}{lst}{}
\floatname{listing}{Listing}
\title{Exploring Vacant Classes in Label-Skewed Federated Learning}
\author{
    Kuangpu Guo\textsuperscript{\rm 1, \rm 3},
    Yuhe Ding\textsuperscript{\rm 2, \rm 3},
    Jian Liang\textsuperscript{\rm 3, \rm 4} \thanks{Corresponding authors.},
    Ran He\textsuperscript{\rm 3, \rm 4},
    Zilei Wang\textsuperscript{\rm 1}, 
    Tieniu Tan\textsuperscript{\rm 5, \rm 4, \rm 3}
}

\affiliations{
    \textsuperscript{\rm 1}University of Science and Technology of China,
    \textsuperscript{\rm 2}Anhui University,\\
    \textsuperscript{\rm 3}NLPR \& MAIS, Institute of Automation, Chinese Academy of Sciences,\\
    \textsuperscript{\rm 4}University of Chinese Academy of Sciences,
    \textsuperscript{\rm 5}Nanjing University\\
    gkp@mail.ustc.edu.cn, \{madao3c, liangjian92\}@gmail.com, zlwang@ustc.edu.cn, \{rhe, tnt\}@nlpr.ia.ac.cn
}

\usepackage{bibentry}

\begin{document}

\maketitle

\begin{abstract}
Label skews, characterized by disparities in local label distribution across clients, pose a significant challenge in federated learning. 
As minority classes suffer from worse accuracy due to overfitting on local imbalanced data, prior methods often incorporate class-balanced learning techniques during local training.
Although these methods improve the mean accuracy across all classes, we observe that vacant classes—referring to categories absent from a client's data distribution—remain poorly recognized.
Besides, there is still a gap in the accuracy of local models on minority classes compared to the global model.
This paper introduces FedVLS, a novel approach to label-skewed federated learning that integrates both vacant-class distillation and logit suppression simultaneously.
Specifically, vacant-class distillation leverages knowledge distillation during local training on each client to retain essential information related to vacant classes from the global model. 
Moreover, logit suppression directly penalizes network logits for non-label classes, effectively addressing misclassifications in minority classes that may be biased toward majority classes.
Extensive experiments validate the efficacy of FedVLS, demonstrating superior performance compared to previous state-of-the-art (SOTA) methods across diverse datasets with varying degrees of label skews. 
Our code is available at https://github.com/krumpguo/FedVLS.
\end{abstract}

\section{Introduction}
Federated learning has emerged as a prominent distributed learning paradigm, lauded for its capability to train a global model without direct access to raw data~\cite{konevcny2016federated, li2020federated1, kairouz2021advances}.
The traditional federated learning (FL) algorithm, FedAvg~\cite{mcmahan2017communication}, follows an iterative process of refining the global model by aggregating parameters from local models, which are initialized with the latest global model parameters and trained across diverse client devices~\cite{sheller2020federated, li2020federated, chai2023survey, luo2022disentangled}.
\begin{figure*}
  \centering
  \includegraphics[width=0.95\textwidth]{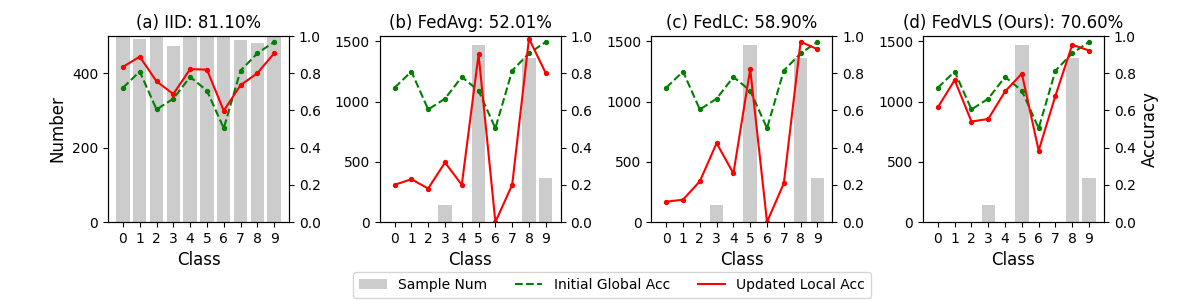}
  \caption{Class-wise accuracy of the initial global model and updated local modelS on IID and label-skewed CIFAR10 data distributions. (a) represents the result updating on IID local data with FedAvg~\cite{mcmahan2017communication}. (b-d) showcase the results updating on skewed data distribution with FedAvg~\cite{mcmahan2017communication}, FedLC~\cite{zhang2022federated}, and our FedVLS, respectively. The value (\%) in each caption corresponds to the accuracy of the global model aggregated from local models.} 
  \label{fig: class-wise acc on client0}
\end{figure*}
In real-world scenarios, local client data often originate from diverse populations or organizations, displaying significant label skews that severely undermine the performance of federated learning ~\cite{hsu2019measuring, yang2021achieving, reguieg2023comparative,zhang2023survey, ye2023heterogeneous}. 

Typically, the local data often consists of majority classes and minority classes, which refer to classes with a large amount of data a small amount of data, respectively~\cite{zhang2022federated, zhang2023survey}.
As evidenced in prior studies \cite{zhang2022federated, chen2022calfat}, the accuracy of minority classes notably decreases after local updates, signaling that the client model is overfitting to local imbalanced data.
Consequently, this results in substantial performance degradation of the global model~\cite {yeganeh2020inverse, li2020federated, liu2022towards}.
To address the issue of lower accuracy in minority classes, previous methods often incorporate class-balanced learning techniques during local training~\cite{zhang2022federated, chen2022calfat, wang2023federated, shen2023federated}. 
Some works~\cite{zhang2022federated, chen2022calfat, shen2023federated, wang2023federated} advocate for calibrating logits according to the client data distribution to balance minority and majority classes. 
However, previous methods have overlooked vacant classes, which refer to classes without data but have highly versatile applications. For instance, in landmark detection~\cite{weyand2020google}, most contributors possess only a subset of landmark categories from places they have lived or traveled. More importantly, these vacant classes can significantly compromise the model's performance, particularly in scenarios with highly skewed label distributions. 

For example, we compare the class-wise accuracy of the initial global model and updated local models using both the classic method FedAvg~\cite{mcmahan2017communication} and one SOTA method FedLC~\cite{zhang2022federated}.
As depicted in Figure~\ref{fig: class-wise acc on client0} (b) and (c), the updated local model exhibits a notable decline in accuracy for vacant classes (e.g., categories 0, 1, 2, 4, 6, and 7) compared to the class-wise accuracy of the initial global model. In extreme cases, the accuracy even decreases close to zero (e.g., category 6).
We posit this severe decline is attributed to the loss of information about vacant classes in the updated local models.
By the way, although FedLC~\cite{zhang2022federated} partially alleviates the performance decline for minority classes, particularly in classes 3, a substantial gap remains compared to the global model.
These results indicate that ignoring vacant classes can lead to a sharp decline in accuracy for those classes, and previous methods still often misclassify minority classes.


Based on these findings, we believe improving the accuracy of vacant and minority classes is crucial for addressing the challenges posed by label skews. 
Therefore, we present FedVLS, a novel approach comprising two pivotal components: vacant-class distillation and logit suppression.
The vacant-class distillation aims to address the performance decline related to vacant classes by distilling vital information from the global model for each client during local training.
Additionally, FedVLS incorporates logit suppression, which regulates the output logit for non-label classes. 
This process emphasizes minimizing the predicted logit values linked to the majority class when handling minority samples, amplifying the penalty for the misclassification of minority classes.
As shown in Figure~\ref{fig: class-wise acc on client0} (d), FedVLS significantly mitigates the decline in accuracy in both vacant and minority classes of the updated local model. 
Consequently, FedVLS effectively reduces overfitting in client models, leading to a significant improvement in the global model's performance.
Our experiments demonstrate that FedVLS consistently outperforms current SOTA federated learning methods across various settings. 
Our contributions are summarized as follows:    
\begin{itemize}
\item We find that prior federated learning methods suffer from vacant classes and propose FedVLS to distill vacant-class-aware knowledge from the global model.
\item FedVLS further presents a logit suppression strategy to address the misclassification of the minority classes, thereby enhancing the generalization of local models.
\item Extensive results validate the effectiveness of both components in FedVLS, outperforming previous state-of-the-art methods across diverse datasets and different degrees of label skews.
\end{itemize}

\section{Related Work}
\subsection{Heterogeneous Federated Learning}
Federated learning faces a significant challenge known as data heterogeneity, also referred to as non-identical and independently distributed (Non-IID) data~\cite{kairouz2021advances, luo2021no, shi2023understanding, guo2024dynamic, guo2024addressing}. 
This challenge encompasses issues such as label skews and domain shifts. 
In this paper, we primarily focus on addressing label skews.
The classic federated learning algorithm, FedAvg~\cite{mcmahan2017communication}, experiences a significant decline in performance when dealing with label skews~\cite{li2019convergence, acar2021federated, luo2024federated}. 
Numerous studies have aimed to mitigate the adverse impacts of label skews. 
For instance, FedProx~\cite{li2020federated} employs a proximal term and SCAFFOLD~\cite{karimireddy2020scaffold} uses a variance reduction approach to constrain the update direction of local models.
Additionally, MOON~\cite{li2021model} and FedProc~\cite{mu2023fedproc} utilize contrastive loss to enhance the agreement between local models and the global model. 
Furthermore, FedConcat~\cite{diao2024exploiting} propose model concatenation, FedMR~\cite{fan2023federated} proposes a manifold reshaping approach, FedGELA~\cite{fan2024federated} uses simplex Equiangular Tight to initialize the local classifier and FedGF~\cite{leerethinking} refine the flat minima searching to alleviate the label skews.
However, these methods often fail to address the issue of vacant classes in highly skewed scenarios. 
We propose FedVLS to effectively mitigate the decline in class-wise accuracy of vacant classes.

\subsection{Learning from Imbalanced Data}
Imbalanced data distribution is pervasive in real-world scenarios, and numerous methods have been proposed to address its impact on model performance~\cite{cui2019class, menon2020long, tan2020equalization, li2022federated, Ma_2023_CVPR}. 
Existing approaches generally fall into two categories: re-weighting~\cite{ cui2019class} and logit-adjustment~\cite{menon2020long, tan2020equalization}.
However, previous works primarily discuss scenarios with long-tailed distributions~\cite{zeng2023global, xiao2023fed}.
These methods may not be directly applicable in federated learning due to the diversity of client data distributions. 
In federated learning, FedLC~\cite{zhang2022federated} and Calfat~\cite{chen2022calfat} introduce logit calibration based on the local data distribution to balance the majority and minority classes. 
FedLMD~\cite{lu2023federated} proposes distillation masks to preserve the information of minority class.
However, they often neglect vacant classes and cannot effectively handle the accuracy decrease of the minority class.
Our method, on the other hand, addresses both existence of vacant classes and the class imbalance between majority and minority classes, making it more practical for real-world scenarios.

\subsection{Knowledge Distillation in Federated Learning}
Knowledge Distillation (KD) has been introduced to federated learning to address issues arising from variations in data distributions and model constructions across clients ~\cite{jeong2018communication,itahara2021distillation,wu2023survey}. 
FedDF~\cite{lin2020ensemble} and FedMD~\cite{li2019fedmd} leverage KD to transfer the knowledge from multiple local models to the global model. 
However, these KD methods typically require a public dataset available to all clients on the server, which presents potential practical challenges. 
Recent methods, such as FEDGEN~\cite{zhu2021data}, DaFKD~\cite{wang2023dafkd}, and DFRD~\cite{luo2023dfrd}, propose training a generator on the server or client to enable data-free federated knowledge distillation. 
However, training the generator adds computational complexity and can often be unstable in cases of extreme label skews~\cite{wu2023survey}. 
Additionally, FedNTD~\cite{lee2022preservation} conducts local-side distillation only for not-true labels to prevent overfitting, while FedHKD~\cite{chen2023best} performs local-side distillation on both logits and class prototypes to align the global and local optimization directions. 
However, these methods perform knowledge distillation across all classes, which may limit the retention of local models' information about vacant classes.
Our method, in contrast, applies knowledge distillation exclusively to vacant classes, preserving vital information about these categories without impacting the learning of other categories or introducing significant computational overhead.

\section{Method}
\subsection{Preliminaries}
In federated learning, we consider a scenario with $N$ clients, where $\mathcal{D}_i$ represents the local training data of client $i$. The combined data $\mathcal{D}=\bigcup_{i=1}^{N} \mathcal{D}_i$ comprises the local data from all clients. These data distributions might differ across clients, encompassing situations where the local training data of some clients only contain samples from a subset of all classes. The overarching goal is to address the optimization problem as follows~\cite{mcmahan2017communication}:
\begin{equation}
\underset{\boldsymbol{\mathcal{\omega}}}{\min}\left[{\mathcal{L}(\bf{\boldsymbol{\omega}})}\overset{def}{=}\sum_{i=1}^{N} {\frac {|\mathcal{D}_i|} {|\mathcal{D}|} {\mathcal{L}_i} (\bf{\boldsymbol{\omega}})} \right],
\end{equation}
where $\mathcal{L}_{i}({\bf \boldsymbol{\omega}})= \mathbb{E}_{(x,y) \sim \mathcal{D}_i} [{\ell}_i (f (x;{\bf \boldsymbol{\omega}}), y)]$ is the empirical loss of the $i$-th client. 
$f(x;{\bf \boldsymbol{\omega}})$ is the output of the model when the input $x$ and model parameter ${\bf \boldsymbol{\omega}}$ are given, and ${\ell}_i$ is the loss function of the $i$-th client. $|\mathcal{D}_i|$ is the number of samples on $\mathcal{D}_i$, $|\mathcal{D}|$ is the number of samples on $\mathcal{D}$. Here, FL expects to learn a global model that can perform well on the entire data $\mathcal{D}$. 

\subsection{Motivation}
\label{subsection: motivation}
When the local training data $\{\mathcal{D}_\textit{i}\}^N_{i=1}$ exhibit label skews, as illustrated in Figure 3 of the technical appendix, there are variations in the quantity of data for the same category across different clients, leading the client models to excessively fit their respective local data distributions. 
This overfitting phenomenon causes divergence during model aggregation, subsequently resulting in inferior global performance~\cite{yeganeh2020inverse, li2020federated, liu2022towards}. 
To address the imbalance between minority and majority classes, previous methodologies~\cite{zhang2022federated, shen2023federated} suggest calibrating logits based on the local data distribution, outlined as follows:
\begin{equation}
\mathcal{L}_{\bf cal} = -\mathbb{E}_{(x,y) \sim \mathcal{D}_i} \log \left(\frac{p(y) \cdot e^{f(x;{\bf\boldsymbol{\omega}})\left[\it{y}\right]}}{\sum_{c}{p(c) \cdot e^{f(x;{\bf\boldsymbol{\omega}})[\it{c}]}}}\right),
\end{equation} 
where $p(y)$ signifies the probability of class $y$ occurring within the client's data distribution, while $f(x;{\bf\boldsymbol{\omega}})[c]$ denotes the logit output for the c-th category.
The calibration technique weights the outputs for all classes in the denominator by $p(c)$. 
However, the probability $p(m)$ for the vacant class in the client data distribution equates to zero, leading to the weighting term for the vacant category, ${p(m) \cdot e^{f(x;{\bf\boldsymbol{\omega}})[m]}}$, also becoming zero.
Consequently, local models prioritize learning the majority and minority classes, gradually disregarding information associated with the vacant categories during local training. 
This gradual shift causes the updated direction of local models to deviate from that of the global model over time.
It's crucial to acknowledge that treating the vacant class merely as a unique minority class is insufficient, an oversight prevalent in prior methodologies.
We assert this drawback significantly contributes to severe instances of local overfitting. 

\begin{figure}[ht]
  \centering
  \includegraphics[width=0.48\textwidth]{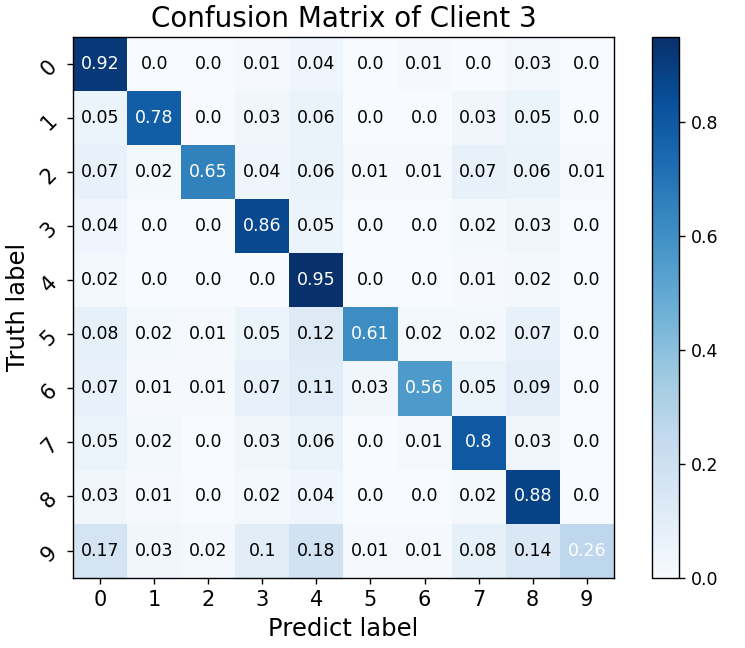}
  \caption{Confusion matrix of client 3 on CIFAR10 dataset with Dirichlet-based label skews ($\beta$ = 0.5) using FedLC~\cite{zhang2022federated}.}
  \label{fig: confusion}
\end{figure}

Our empirical observations reveal a substantial decrease in class-wise accuracy for vacant classes (such as categories 0, 1, 2, 4, 6, and 7) after the local update, as illustrated in Figure~\ref{fig: class-wise acc on client0} (b) and (c). 
Notably, in specific cases (such as category 6), this accuracy even drops close to zero.
The specific experimental setup and other analyses can be found in the technical appendix.
By the way, we find the updated class-wise accuracy for the minority classes (e.g., category 3) continues to display a notable decline, maintaining a significant gap compared to the IID scenario.  
Through the analysis of the confusion matrix in Figure~\ref{fig: confusion}, we find that vacant and minority classes are still frequently misclassified as majority classes.
Thus, we aim to develop different objectives to alleviate these two issues, respectively.      

\subsection{Vacant-class Distillation}
\label{subsection: Vacant-class Distillation}
Motivated by the above observations and analyses of vacant classes, we propose to prevent the disappearance of information related to vacant classes during local training.
The global model harbors valuable insights, particularly regarding the prediction of vacant classes, making it an exceptional teacher for each client. 
Hence, we introduce vacant-class distillation, aimed at preserving the global perspective of vacant classes for clients through knowledge distillation. To achieve this, we utilize the Kullback-Leibler Divergence loss function, as outlined below:
\begin{equation}
\mathcal{L}_{\bf dis} = \mathbb{E}_{(x,y) \sim \mathcal{D}_i} \sum_{o \in \mathbb{O}} {q^g(o;x) \log \left[ \frac{q(o;x)}{q^g(o;x)} \right]},
\label{eq: dist}
\end{equation} 
\begin{equation}
\text{where}\; q(o;x)= \frac{\exp(f(x;{\bf\boldsymbol{\omega}})[o])}{\sum_{c \in \mathbb{O}} {\exp(f(x;{\bf\boldsymbol{\omega}})[c])}}
\notag
\end{equation} 
denotes the output for the $o$-th class of the local model using softmax within the vacant classes, and $q^g(o;x) = \frac{\exp(f(x;{\bf\boldsymbol{\omega}^g})[o])}{\sum_{c \in \mathbb{O}}{\exp(f(\it{x};{\bf\boldsymbol{\omega}^g})[c]}}$ denotes the same for the global model.
$\mathbb{O}$ represents the set that contains all vacant classes within the local client data and $\boldsymbol{\omega}^g$ denotes the parameters of the global model. 

Unlike FedNTD~\cite{lee2022preservation}, which encourages the client model to closely match the global model's output for not-true labels, thereby limiting the knowledge protection for vacant class, our loss function ensures that the local model replicates the global model's outputs only for vacant-class labels. 
This approach preserves the predictive capability for vacant categories significantly.
Moreover, the computational overhead introduced by this loss function is minimal, enhancing its practical implementation.
Additional comparisons with other distillation-based methods and further analyses are provided in the technical appendix.

\subsection{Logit Suppression}
\label{subsection: Logit Suppression} 
Previous methods still often suffer from low accuracy in the minority classes of local models, a factor that requires mitigation to enhance the generalization capabilities of these models. 
To identify the root cause of this issue, we analyzed the confusion matrix of the local model in client 3 on the entire test dataset using FedLC~\cite{zhang2022federated}, where the training data distribution is shown in the fourth column of Figure 3 (a) in the technical appendix. 
As shown in Figure~\ref{fig: confusion}, minority classes (e.g., categories 2, 5, and 6) are frequently misclassified as majority classes (e.g., categories 0, 4, and 8). 
It is evident that, in the model's output for minority samples, the majority class tends to have a higher logit value, leading to the misclassification of  minority classes.
Therefore, we implement regularization on non-label class logits to penalize the majority class output for  minority samples. 
To avoid non-trivial optimization over direct logits, we aim to minimize the following objective for each class:
\begin{equation}
\mathcal{L}^{c}_{\bf logit} = \log \left(\mathbb{E}_{(x,y) \sim \mathcal{D}_i} \mathbb{I}({y \neq c}) \cdot e^{f(x;\boldsymbol{\omega}) [c]}\right), 
\end{equation}
where $\mathbb{I}$ is an indicator function with value 1 when ${y \neq c}$. 
We use the log function to increase the proportion of loss values for minority categories. 
Since minority samples are prone to be more frequently misclassified into majority classes, the higher weight should be assigned to the logits of majority categories in non-labeled outputs. 
Therefore, we weight the loss function $\mathcal{L}^{c}_{\bf logit}$ using the probability of occurrence $p(c)$ for each class as follows: 
\begin{equation}
\mathcal{L}_{\bf logit} = \sum_{c} p(c) \cdot \mathcal{L}^{c}_{\bf logit}.
\end{equation}  
This adaptation prompts the learning process to pay more attention to the penalty for incorrectly classifying minority class samples as majority classes.
As a result, the model is encouraged to refine its prediction across diverse classes, thereby improving its overall generalization capability.

\begin{table*}[ht]
\centering
    \caption{Performance overview for different degrees of Dirichlet-based label skews. All results are (re)produced by us and are averaged over 3 runs (mean ± std). \textbf{Bold} is the best result, \underline{underline} is the second-best.}
\label{tab: accuracy Dir}
\resizebox{1\textwidth}{!}{ 
\begin{tabular}{>{\large}l>{\large}c>{\large}c>{\large}c>{\large}c>{\large}c>{\large}c>{\large}c>{\large}c>{\large}c>{\large}c>{\large}c>{\large}c>{\large}c}
\toprule
\multirow{2}{*}{Method(venue)}  & \multicolumn{3}{c}{\textbf{MNIST}} & \multicolumn{3}{c}{\textbf{CIFAR10}} & \multicolumn{3}{c}{\textbf{CIFAR100}} & \multicolumn{3}{c}{\textbf{TinyImageNet}} \\ \cmidrule(lr){2-4}  \cmidrule(lr){5-7} \cmidrule(lr){8-10} \cmidrule(lr){11-13}
    & $\beta=0.5$ & $\beta=0.1$ & $\beta=0.05$ & $\beta=0.5$ & $\beta=0.1$ & $\beta=0.05$ & $\beta=0.5$ & $\beta=0.1$ & $\beta=0.05$ & $\beta=0.5$ & $\beta=0.1$ & $\beta=0.05$   \\ \midrule
FedAvg (AISTATS 2017) & 98.96\scalebox{1.25}{\scriptsize$\pm0.00$} & 96.69\scalebox{1.25}{\scriptsize$\pm0.00$} & 94.77\scalebox{1.25}{\scriptsize$\pm0.44$} & 91.46\scalebox{1.25}{\scriptsize$\pm0.55$} & 82.00\scalebox{1.25}{\scriptsize$\pm0.75$} & 62.90\scalebox{1.25}{\scriptsize$\pm0.95$} & 72.22\scalebox{1.25}{\scriptsize$\pm0.34$} & 66.18\scalebox{1.25}{\scriptsize$\pm0.35$} & 62.13\scalebox{1.25}{\scriptsize$\pm0.09$} & 47.02\scalebox{1.25}{\scriptsize$\pm0.40$} & 39.90\scalebox{1.25}{\scriptsize$\pm0.27$} & 35.21\scalebox{1.25}{\scriptsize$\pm0.47$} \\[4pt]
FedProx (MLSys 2020) & 98.93\scalebox{1.25}{\scriptsize$\pm0.00$} & 96.42\scalebox{1.25}{\scriptsize$\pm0.00$} & 94.95\scalebox{1.25}{\scriptsize$\pm0.24$} & 92.24\scalebox{1.25}{\scriptsize$\pm0.78$} & 82.65\scalebox{1.25}{\scriptsize$\pm1.33$} & 63.14\scalebox{1.25}{\scriptsize$\pm0.41$} & 72.65\scalebox{1.25}{\scriptsize$\pm0.60$} & 66.61\scalebox{1.25}{\scriptsize$\pm0.22$} & 62.23\scalebox{1.25}{\scriptsize$\pm0.20$} & 45.76\scalebox{1.25}{\scriptsize$\pm0.50$} & 40.26\scalebox{1.25}{\scriptsize$\pm0.51$} & 35.22\scalebox{1.25}{\scriptsize$\pm0.17$} \\[4pt]
MOON (CVPR 2021) & 99.18\scalebox{1.25}{\scriptsize$\pm0.01$} & 96.94\scalebox{1.25}{\scriptsize$\pm0.12$} & 93.39\scalebox{1.25}{\scriptsize$\pm0.21$} & 92.13\scalebox{1.25}{\scriptsize$\pm0.35$} & 83.38\scalebox{1.25}{\scriptsize$\pm0.43$} & 61.34\scalebox{1.25}{\scriptsize$\pm0.77$} & 72.87\scalebox{1.25}{\scriptsize$\pm0.1 1$} & 66.12\scalebox{1.25}{\scriptsize$\pm0.32$} & 60.45\scalebox{1.25}{\scriptsize$\pm0.41$} & 42.26\scalebox{1.25}{\scriptsize$\pm0.36$} & 36.88\scalebox{1.25}{\scriptsize$\pm0.53$} & 33.61\scalebox{1.25}{\scriptsize$\pm0.35$} \\[4pt]
FedEXP (ICLR 2023) & 97.57\scalebox{1.25}{\scriptsize$\pm0.49$} & 91.59\scalebox{1.25}{\scriptsize$\pm0.48$} & 92.54\scalebox{1.25}{\scriptsize$\pm1.08$} & 92.31\scalebox{1.25}{\scriptsize$\pm0.52$} & 83.48\scalebox{1.25}{\scriptsize$\pm1.15$} & 63.22\scalebox{1.25}{\scriptsize$\pm0.51$} & 72.41\scalebox{1.25}{\scriptsize$\pm0.39$} & 66.74\scalebox{1.25}{\scriptsize$\pm0.19$} & 62.24\scalebox{1.25}{\scriptsize$\pm0.18$} & 47.00\scalebox{1.25}{\scriptsize$\pm0.23$} & 40.58\scalebox{1.25}{\scriptsize$\pm0.15$} & 34.95\scalebox{1.25}{\scriptsize$\pm0.18$} \\[4pt]
FedLC (ICML 2022) & 98.97\scalebox{1.25}{\scriptsize$\pm0.01$} & 95.59\scalebox{1.25}{\scriptsize$\pm0.05$} & 85.56\scalebox{1.25}{\scriptsize$\pm0.18$} & 91.98\scalebox{1.25}{\scriptsize$\pm0.63$} & 82.24\scalebox{1.25}{\scriptsize$\pm0.53$} & 57.31\scalebox{1.25}{\scriptsize$\pm0.97$} & 72.69\scalebox{1.25}{\scriptsize$\pm0.30$} & 66.20\scalebox{1.25}{\scriptsize$\pm0.20$} & 59.18\scalebox{1.25}{\scriptsize$\pm0.11$} & 48.01\scalebox{1.25}{\scriptsize$\pm0.21$} & 41.46\scalebox{1.25}{\scriptsize$\pm0.37$} & 35.56\scalebox{1.25}{\scriptsize$\pm0.58$} \\[4pt]
FedRS (KDD 2021) & 99.03\scalebox{1.25}{\scriptsize$\pm0.00$} & 96.67\scalebox{1.25}{\scriptsize$\pm0.01$} & 94.60\scalebox{1.25}{\scriptsize$\pm0.40$} & \underline{92.55}\scalebox{1.25}{\scriptsize$\pm0.68$} & \underline{83.95}\scalebox{1.25}{\scriptsize$\pm0.35$} & 63.17\scalebox{1.25}{\scriptsize$\pm0.57$} & 72.99\scalebox{1.25}{\scriptsize$\pm0.20$} & 66.84\scalebox{1.25}{\scriptsize$\pm0.25$} & 62.19\scalebox{1.25}{\scriptsize$\pm0.06$} & 47.95\scalebox{1.25}{\scriptsize$\pm0.43$} & 41.77\scalebox{1.25}{\scriptsize$\pm0.25$} & 35.82\scalebox{1.25}{\scriptsize$\pm0.20$} \\[4pt]
FedSAM (ICML2022) & 99.21\scalebox{1.25}{\scriptsize$\pm0.00$} & \underline{97.24}\scalebox{1.25}{\scriptsize$\pm0.00$} & 95.17\scalebox{1.25}{\scriptsize$\pm0.42$} & 92.37\scalebox{1.25}{\scriptsize$\pm1.33$} & 81.19\scalebox{1.25}{\scriptsize$\pm0.32$} & 63.11\scalebox{1.25}{\scriptsize$\pm1.05$ }& 72.96\scalebox{1.25}{\scriptsize$\pm0.25$} & 67.50\scalebox{1.25}{\scriptsize$\pm0.19$} & 61.32\scalebox{1.25}{\scriptsize$\pm0.14$} & 48.43\scalebox{1.25}{\scriptsize$\pm1.42$} & 43.96\scalebox{1.25}{\scriptsize$\pm1.02$} & 41.14\scalebox{1.25}{\scriptsize$\pm0.23$} \\[4pt]
FedNTD (NeurIPS 2022) & 99.15\scalebox{1.25}{\scriptsize$\pm0.04$} & 96.67\scalebox{1.25}{\scriptsize$\pm0.17$} & 94.30\scalebox{1.25}{\scriptsize$\pm0.71$} & 92.46\scalebox{1.25}{\scriptsize$\pm0.19$} & 83.23\scalebox{1.25}{\scriptsize$\pm0.22$} & 68.71\scalebox{1.25}{\scriptsize$\pm0.27$} & \underline{73.43}\scalebox{1.25}{\scriptsize$\pm0.15$} & 68.00\scalebox{1.25}{\scriptsize$\pm0.50$} & 63.71\scalebox{1.25}{\scriptsize$\pm0.19$} & 48.02\scalebox{1.25}{\scriptsize$\pm1.05$} & 45.11\scalebox{1.25}{\scriptsize$\pm0.21$} & 40.65\scalebox{1.25}{\scriptsize$\pm0.26$} \\[4pt]
FedMR (TMLR 2023) & 98.95\scalebox{1.25}{\scriptsize$\pm0.02$} & 96.73\scalebox{1.25}{\scriptsize$\pm0.08$} & 95.34\scalebox{1.25}{\scriptsize$\pm0.50$} & 91.98\scalebox{1.25}{\scriptsize$\pm0.55$} & 82.09\scalebox{1.25}{\scriptsize$\pm0.42$} & 63.54\scalebox{1.25}{\scriptsize$\pm0.69$} & 71.94\scalebox{1.25}{\scriptsize$\pm0.36$} & 67.57\scalebox{1.25}{\scriptsize$\pm0.37$} & 63.75\scalebox{1.25}{\scriptsize$\pm0.24$} & 47.21\scalebox{1.25}{\scriptsize$\pm0.53$} & 40.35\scalebox{1.25}{\scriptsize$\pm0.26$} & 35.94\scalebox{1.25}{\scriptsize$\pm0.46$} \\[4pt]
FedLMD (MM 2023) & 99.17\scalebox{1.25}{\scriptsize$\pm0.03$} & 97.18\scalebox{1.25}{\scriptsize$\pm0.12$} & 95.33\scalebox{1.25}{\scriptsize$\pm0.53$} & 92.50\scalebox{1.25}{\scriptsize$\pm0.34$} & 83.14\scalebox{1.25}{\scriptsize$\pm0.19$} & \underline{70.50}\scalebox{1.25}{\scriptsize$\pm0.29$} & 73.30\scalebox{1.25}{\scriptsize$\pm0.30$} & \underline{68.83}\scalebox{1.25}{\scriptsize$\pm0.35$} & 64.10\scalebox{1.25}{\scriptsize$\pm0.19$} & 48.43\scalebox{1.25}{\scriptsize$\pm0.48$} & 44.03\scalebox{1.25}{\scriptsize$\pm0.25$} & 41.18\scalebox{1.25}{\scriptsize$\pm0.27$} \\[4pt]
FedConcat (AAAI 2024) & 99.04\scalebox{1.25}{\scriptsize$\pm0.01$} & 96.99\scalebox{1.25}{\scriptsize$\pm0.11$} & 95.02\scalebox{1.25}{\scriptsize$\pm0.47$} & 92.45\scalebox{1.25}{\scriptsize$\pm0.29$} & 82.83\scalebox{1.25}{\scriptsize$\pm0.21$} & 64.30\scalebox{1.25}{\scriptsize$\pm0.28$} & 73.27\scalebox{1.25}{\scriptsize$\pm0.28$} & 68.57\scalebox{1.25}{\scriptsize$\pm0.34$} & 63.74\scalebox{1.25}{\scriptsize$\pm0.13$} & 48.45\scalebox{1.25}{\scriptsize$\pm0.44$} & 47.32\scalebox{1.25}{\scriptsize$\pm0.21$} & 43.44\scalebox{1.25}{\scriptsize$\pm0.21$} \\[4pt]
FedGF (ICML 2024) & \underline{99.22}\scalebox{1.25}{\scriptsize$\pm0.00$} & \textbf{97.35}\scalebox{1.25}{\scriptsize$\pm0.00$} & \underline{95.36}\scalebox{1.25}{\scriptsize$\pm0.28$} & 92.52\scalebox{1.25}{\scriptsize$\pm0.22$} & 82.91\scalebox{1.25}{\scriptsize$\pm0.16$} & 69.61\scalebox{1.25}{\scriptsize$\pm0.47$} & 73.30\scalebox{1.25}{\scriptsize$\pm0.25$} & 68.70\scalebox{1.25}{\scriptsize$\pm0.20$} & \underline{64.48}\scalebox{1.25}{\scriptsize$\pm0.08$} & \underline{48.52}\scalebox{1.25}{\scriptsize$\pm0.23$} & \underline{47.64}\scalebox{1.25}{\scriptsize$\pm0.16$} & \underline{44.71}\scalebox{1.25}{\scriptsize$\pm0.20$} \\ \cmidrule{1-13}
\rowcolor[gray]{0.9} 
\textbf{FedVLS (Ours)} & \textbf{99.23}\scalebox{1.25}{\scriptsize$\pm0.00$} & \underline{97.24}\scalebox{1.25}{\scriptsize$\pm0.00$} & \textbf{95.56}\scalebox{1.25}{\scriptsize$\pm0.12$} & \textbf{92.66}\scalebox{1.25}{\scriptsize$\pm0.14$} & \textbf{84.35}\scalebox{1.25}{\scriptsize$\pm0.04$} & \textbf{75.71}\scalebox{1.25}{\scriptsize$\pm0.28$} & \textbf{73.49}\scalebox{1.25}{\scriptsize$\pm0.80$} & \textbf{69.02}\scalebox{1.25}{\scriptsize$\pm0.18$} & \textbf{65.71}\scalebox{1.25}{\scriptsize$\pm0.01$} & \textbf{48.54}\scalebox{1.25}{\scriptsize$\pm0.12$} & \textbf{47.73}\scalebox{1.25}{\scriptsize$\pm0.13$} & \textbf{45.23}\scalebox{1.25}{\scriptsize$\pm0.15$} \\ \bottomrule
\end{tabular} 
}
\end{table*}

\begin{table}
\centering
\caption{Performance overview for quantity-based label skews. $s$ presents the number of shards per client.}
\label{tab: accuracy quantity}
\resizebox{0.48\textwidth}{!}{ 
\begin{tabular}{lccc}
\toprule
\multirow{2}{*}{Method(venue)} & \textbf{CIFAR10} & \textbf{CIFAR100} & \textbf{TinyImageNet}  \\ \cmidrule{2-4} 
    & $s=2$ & $s=20$ & $s=40$ \\ \midrule
FedAvg (AISTATS 2017) & 44.63\scriptsize$\pm0.77$ & 63.14\scriptsize$\pm0.03$ & 30.28\scriptsize$\pm0.12$ \\
FedProx (MLSys 2020) & 48.65\scriptsize$\pm0.59$ & 62.10\scriptsize$\pm0.10$ & 28.14\scriptsize$\pm0.93$ \\
MOON (CVPR 2021) & 38.24\scriptsize$\pm1.00$ & 57.33\scriptsize$\pm0.06$ & 26.25\scriptsize$\pm0.73$ \\
FedEXP (ICLR 2023) & 41.11\scriptsize$\pm0.26$ & 62.61\scriptsize$\pm0.06$ & 29.38\scriptsize$\pm0.19$ \\
FedLC (ICML 2022) & 55.14\scriptsize$\pm0.26$ & 61.56\scriptsize$\pm0.03$ & 26.29\scriptsize$\pm1.00$ \\
FedRS (KDD 2021) & 42.20\scriptsize$\pm1.49$ & 61.53\scriptsize$\pm0.03$ & 28.31\scriptsize$\pm0.06$ \\
FedSAM (ICML2022) & 36.97\scriptsize$\pm1.18$ & 63.50\scriptsize$\pm0.01$ & 37.55\scriptsize$\pm0.10$ \\
FedNTD (NeurIPS 2022) & {67.35}\scriptsize$\pm0.19$ & 63.74\scriptsize$\pm0.01$ & 37.19\scriptsize$\pm0.07$ \\ 
FedMR (TMLR 2023) & 46.55\scriptsize$\pm0.64$ & 63.55\scriptsize$\pm0.03$ & 28.45\scriptsize$\pm0.10$ \\ 
FedLMD (MM 2023) & \textbf{68.52}\scriptsize$\pm0.34$ & 63.51\scriptsize$\pm0.02$ & 32.29\scriptsize$\pm0.08$ \\
FedConcat (AAAI 2024) & 62.00\scriptsize$\pm0.28$ & 63.87\scriptsize$\pm0.01$ & 42.95\scriptsize$\pm0.06$ \\
FedGF (ICML 2024) & 66.97\scriptsize$\pm0.45$ & \underline{63.90}\scriptsize$\pm0.02$ & \underline{43.55}\scriptsize$\pm0.06$ \\\cmidrule{1-4} \rowcolor[gray]{0.9}
\textbf{FedVLS (Ours)} & \underline{68.03}\scriptsize$\pm0.$18 & \textbf{64.95}\scriptsize$\pm0.01$ & \textbf{43.97}\scriptsize$\pm0.04$ \\
\bottomrule
\end{tabular}
}
\end{table}

\subsection{Overall Objective}
As of now, we have elaborated extensively on our strategy to tackle the problem of loss of information about vacant classes in previous methods through knowledge distillation.
Moreover, we mitigate the decrease in minority classes of the updated local model by regulating non-label logits directly, to further alleviate local overfitting issues. 
In summary, we propose the comprehensive method named FedVLS, whose objective is as follows:
\begin{equation}
\mathcal{L}({\bf \boldsymbol{\omega}}) = \mathcal{L}_{\bf cal} + \lambda \cdot \mathcal{L}_{\bf dis} + \mathcal{L}_{\bf logit},
\label{eq: final loss}
\end{equation}  
where $\lambda$ is a non-negative hyperparameter to control the contribution of vacant-class distillation. 
In the loss function of our FedVLS, we attain new knowledge from the observed class in local data distribution using the $\mathcal{L}_{\bf cal}$ and $\mathcal{L}_{\bf logit}$. 
In the meanwhile, we preserve the previous knowledge on the vacant classes by following the global model’s perspective using the $\mathcal{L}_{\bf dis}$. 
By combining vacant-class distillation and logit suppression, FedVLS can effectively manipulate various levels of label skews.
Algorithm 1 in the technical appendix shows the overflow of our method.

\section{Experiments}
\label{sec: experiment}

\subsection{Setups}
\label{Setups}
\textbf{Datasets}
\label{Datasets}
We evaluate the effectiveness of our approach across various image classification datasets, including MNIST~\cite{deng2012mnist}, CIFAR10~\cite{krizhevsky2009learning}, CIFAR100~\cite{krizhevsky2009learning}, and TinyImageNet~\cite{le2015tiny}. 
We partitioned each dataset into distinct training and test sets. 
Subsequently, the training set undergoes further division into non-overlapping subsets, distributed among different clients. 
The global model's performance is then assessed on the test set. 
We follow the settings outlined in~\cite{li2022federated} and introduce two prevalent forms of label skews: Dirichlet-based and quantity-based. 
In the quantity-based label skews, all training data is grouped by label and allocated into shards with imbalanced quantities. 
The parameter $s$ signifies the number of shards per client, regulating the level of label skews~\cite{lee2022preservation}. 
In the Dirichlet-based label skews, clients receive samples for each class based on the Dirichlet distribution~\cite{zhu2021federated}, denoted as $D(\beta)$. 
Here, the parameter $\beta$ controls the degree of label skews, with lower values indicating higher label skews. 
Notably, each client's training data may encompass majority classes, minority classes, and even vacant classes, which is more practical. 

\textbf{Models and baselines}
Following a prior study~\cite{shi2022towards}, our primary network architecture for all experiments, except MNIST, predominantly relies on MobileNetV2~\cite{sandler2018mobilenetv2}. 
For the MNIST, we adopt a deep neural network (DNN) containing three fully connected layers as the backbone.
Our baseline models encompass conventional approaches to tackle data heterogeneity issues, including FedProx~\cite{li2020federated}, MOON~\cite{li2021model}, FedSAM~\cite{qu2022generalized}, FedEXP~\cite{jhunjhunwala2023fedexp}, FedConcat~\cite{diao2024exploiting}, and FedGF~\cite{leerethinking}. 
To ensure a fair comparison, we also assess our method against FedRS~\cite{li2021fedrs}, FedLC~\cite{zhang2022federated}, FedNTD~\cite{lee2022preservation}and FedLMD~\cite{lu2023federated}, which also focus on addressing label skews in federated learning.

\begin{figure*}[t]
  \centering
  \includegraphics[width=0.9\textwidth]{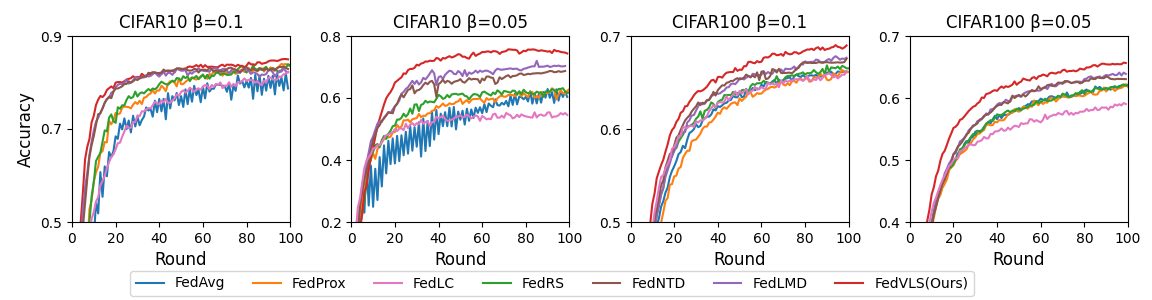}
  \caption{The test accuracy over each communication round during training for different levels of Dirichlet-based label skews ($\beta \in \{0.1, 0.05\}$) on CIFAR10 and CIFAR100 datasets.}
  \label{fig: convergence curve}
\end{figure*}

\textbf{Implementation details} 
\label{Implementation details}
We set the number of clients N to 10 and implement full client participation. 
We run 100 communication rounds for all experiments on the CIFAR10/100 datasets and 50 communication rounds on the MNIST and TinyImageNet datasets.
Within each communication round, local training spans 5 epochs for MNIST and 10 epochs for the other datasets.
For FedConcat~\cite{diao2024exploiting} and FedGF~\cite{leerethinking}, we followed the original paper's settings for communication rounds and local epochs.
We employ stochastic gradient descent (SGD) optimization with a learning rate of 0.01, a momentum of 0.9, and a batch size of 64. 
Weight decay is set to $10^{-5}$ for MNIST and CIFAR10 and $10^{-4}$ for CIFAR100 and TinyImageNet.
The hyperparameter $\lambda$ of FedVLS in Equation~\ref{eq: final loss} is set to 0.1 for MNIST and CIFAR10, while it is set to 0.5 for CIFAR100 and TinyImageNet. 
Following pFedMe~\cite{t2020personalized}, we conduct three trials for each experimental setting and report the mean accuracy and standard deviation of the maximum accuracy achieved by the global model during the training process.
More implementation details and experimental results can be found in the technical appendix at https://github.com/krumpguo/FedVLS.

\begin{figure}[ht]
  \centering
  \includegraphics[width=0.46\textwidth]{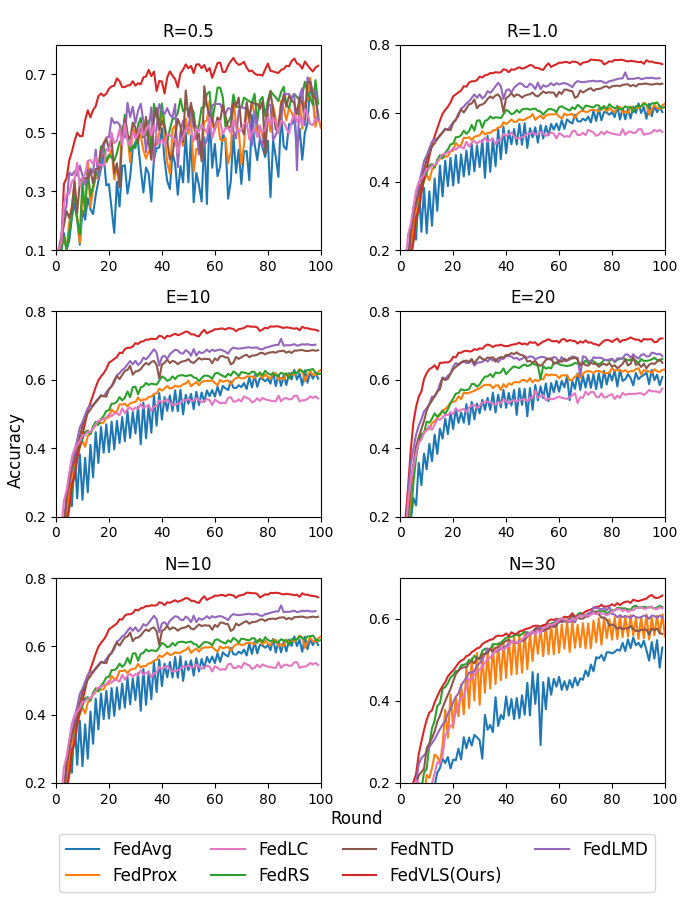}
  \caption{Sensitivity analysis on the client participating rates $\mathbf{R}$, local epochs $\mathbf{E}$, and client numbers $\mathbf{N}$. }
  \label{fig: ablation}
\end{figure}

\subsection{Results}
\textbf{Results under various levels of label skews and datasets} 
Table~\ref{tab: accuracy Dir} presents the performance results of various methods with different levels of Dirichlet-based label skews ($\beta \in \{0.5, 0.1, 0.05\}$).
Our method consistently achieves notably higher accuracy compared to other SOTA methods.
As the degree of label skews increases, competing methods struggle to maintain their performance levels. 
For instance, FedLC~\cite{zhang2022federated} experiences a substantial decline, dropping even below the performance of the classic method FedAvg when $\beta=0.05$. 
This decline stems from each client having numerous vacant classes in extreme cases, a factor overlooked by FedLC~\cite{zhang2022federated}. 
Conversely, our method consistently upholds excellent performance, especially in highly skewed label distribution scenarios. 
For instance, in the case of the CIFAR10 dataset with $\beta=0.05$, our method achieves an impressive test accuracy of 75.71\%, surpassing FedAvg by 12.81\%.
This outcome highlights the efficacy of our approach in addressing the accuracy decline observed in both vacant and minority classes, effectively mitigating instances of overfitting in local data distributions.
Additionally, we present the performance of these methods for quantity-based label distribution skews in Table~\ref{tab: accuracy quantity}, further emphasizing the superiority of our method.

\begin{table}[ht]
\centering
\caption{Results of different methods under various backbones with Dirichlet-based label skews on CIFAR10 dataset.}
\label{tab: accuracy diff backbone}
\resizebox{0.48\textwidth}{!}{ 
\begin{tabular}{lcccccc}
\toprule
\multirow{2}{*}{Method(venue)} & \multicolumn{2}{c}{\textbf{ResNet18}} & \multicolumn{2}{c}{\textbf{ResNet32}} & \multicolumn{2}{c}{\textbf{MobileNetV2}} \\ \cmidrule(lr){2-3}  \cmidrule(lr){4-5} \cmidrule(lr){6-7} 
& $\beta$=0.1 & $\beta$=0.05 & $\beta$=0.1 & $\beta$=0.05 & $\beta$=0.1 & $\beta$=0.05 \\ \midrule
FedAvg (AISTATS 2017) & 73.84 & 58.54 & 79.38 & 55.41 & 82.00 & 62.90 \\
FedProx (MLSys 2020) & 74.68 & 58.14 & 80.60 & 62.51 & 82.65 & 63.14 \\
MOON (CVPR 2021) & 74.04 & 55.41 & 76.91 & 51.85 & 83.38 & 61.34 \\
FedEXP (ICLR 2023) & 72.80 & 58.04 & 78.36 & 53.35 & 83.48 & 63.22 \\
FedLC (ICML 2022)  & 73.15 & 48.94 & 77.71 & 55.41 & 82.24 & 57.31 \\
FedRS (KDD 2021) & 76.38 & 57.47 & \underline{82.03} & 66.87 & \underline{83.95} & 63.17 \\
FedSAM (ICML2022) & 68.42 & 55.42 & 75.66 & 58.88 & 81.19 & 63.11 \\
FedNTD (NeurIPS 2022) & 76.76 & 60.01 & 79.75 & 65.96 & 83.23 & 68.71 \\ 
FedLMD (MM 2023) & \underline{77.02} & \underline{65.80} & 81.76 & \underline{68.04} & 83.14 & \underline{70.50} \\ 
FedConcat (AAAI 2024) & 76.33 & 59.83 & 70.32 & 61.86 & 82.83 & 64.30 \\ 
FedGF (ICML 2024) & 76.74 & 64.44 & 81.44 & 67.83 & 82.91 & 69.61 \\ \cmidrule{1-7} \rowcolor[gray]{0.9}
\textbf{FedVLS (Ours)} & \textbf{78.00} & \textbf{68.33} & \textbf{82.44} & \textbf{68.84} & \textbf{84.35} & \textbf{75.71} \\
\bottomrule
\end{tabular}
}
\end{table}

\textbf{Communication efficiency}
Figure~\ref{fig: convergence curve} illustrates the accuracy over each communication round throughout the training process. 
Our method showcases quicker convergence and higher accuracy when compared to the other six methods. Due to the differences in communication rounds among FedConcat~\cite{diao2024exploiting}, FedGF~\cite{leerethinking} and our approach, we have not included the convergence curves for these two methods. Unlike its counterparts, our approach displays a more consistent upward trend. Moreover, our method exhibits a significant improvement as the skews in the data distribution increase. These outcomes underscore the substantial communication efficiency of our method compared with other approaches.

\subsection{Analysis}
\textbf{Impact of participating rates}
To begin with, we analyze our model's performance against SOTA methods across varying client participation rates. 
Unless specified otherwise, our experiments focus on the CIFAR10 dataset with a Dirichlet-based skew parameter of $\beta=0.05$. 
Initially, we set the client participation rate $R$ within the range $\{0.5, 1.0\}$.
As illustrated in the top row of Figure~\ref{fig: ablation}, our method consistently outperforms other approaches across all participation rates, showcasing a faster convergence rate. 
Notably, as the participation rate decreases, several methods display highly unstable convergence. 
This instability is expected, as a lower client participation rate amplifies the divergence between randomly participating clients and the global model, resulting in erratic convergence.
In contrast, our method exhibits a relatively stable convergence trend, highlighting its robustness to varying participation rates.

\begin{table}[t]
\centering
\caption{Results under different values of hyperparameter $\lambda$ with Dirichlet-based label skews ($\beta=0.05$) on CIFAR10 and CIFAR100 datasets.}
\label{fig: diff beta}
\resizebox{0.48\textwidth}{!}{ 
\begin{tabular}{lcccccc}
\toprule
$\lambda$ & 0.05 & 0.1 & 0.25 & 0.5 & 1 \\ \midrule
\textbf{CIFAR10}  & 74.70 & \textbf{75.71} & 75.47 & 75.29 & 74.98 \\ 
\textbf{CIFAR100}  & 65.49 & 65.57 & 65.63 & \textbf{65.71} & 65.18 \\
\bottomrule
\end{tabular}
}
\end{table}

\begin{table}[t]
\centering
\caption{Effectiveness of each loss function in FedVLS with Dirichlet-based label skews ($\beta=0.05$) on various datasets. (The value) represents the improvement over the first row.}
\label{tab: diff objective}
\resizebox{0.48\textwidth}{!}{ 
\begin{tabular}{cclll}
\toprule
$\mathcal{L}_{\bf dis}$ & $\mathcal{L}_{\bf logit}$ & \textbf{CIFAR10} & \textbf{CIFAR100} & \textbf{TinyImageNet} \\ \midrule
\xmark & \xmark & 57.31 & 59.18 & 35.56 \\ 
\xmark & \cmark & 70.25({\color{red}+12.94}) & 64.24({\color{red}+5.06}) & 39.04({\color{red}+3.48}) \\ 
\cmark & \xmark & 71.53({\color{red}+14.22}) & 65.28({\color{red}+6.10}) & 44.90({\color{red}+9.34}) \\ 
\cmark & \cmark & 75.71({\color{red}+18.40}) & 65.71({\color{red}+6.53}) & 45.23({\color{red}+9.67}) \\ 
\bottomrule
\end{tabular}
}
\end{table}

\textbf{Impact of local epochs}
In this analysis, we investigate variations in the number of local epochs per communication round, represented as $E$, considering values from $\{10, 20\}$. 
An intriguing observation emerges, particularly noticeable when $E$ equals 20: several methods, notably FedNTD~\cite{lee2022preservation}, exhibit declining accuracy in the later stages of training, as depicted in the second row of Figure~\ref{fig: ablation}. 
This decline is attributed to larger $E$ values, making these models more susceptible to overfitting local data distribution as training progresses.
In contrast, our method sustains a consistent and improving performance even with larger $E$ values and consistently outperforms all other methods. 

\textbf{Impact of client numbers}
To underscore the resilience of our method in scenarios involving an increasing number of clients, we divide the CIFAR10 dataset into 10 and 30 clients, showcasing their convergence curves in the final row of Figure~\ref{fig: ablation}. 
Remarkably, our method consistently outperforms the baseline methods, regardless of the number of clients. 
An interesting trend emerges where, with the expanding number of clients, many methods exhibit slower and less stable convergence.
In contrast, FedVLS maintains a consistent trend of rapid and stable convergence across these varied client numbers. 
Additional ablation study results concerning participating rates, local epochs, and the number of clients can be found in the technical appendix.
 
\textbf{Impact of different backbones}
Apart from MobileNetV2, we conduct experiments using ResNet18 and ResNet32. 
The skew parameter, denoted as $\beta$, is set to 0.1 and 0.05. 
The results are presented in Table~\ref{tab: accuracy diff backbone}, demonstrating our method, FedVLS, consistently outperforms the baseline methods.
These experiments underscore the versatility and robustness of FedVLS in real-world federated learning scenarios employing various backbone architectures.

\textbf{Robustness to hyperparameter $\lambda$}
To demonstrate the robustness of our method concerning hyperparameter selection, we conduct experiments using various values of $\lambda$ on the CIFAR10 and CIFAR100 datasets.
The findings, presented in Table~\ref{fig: diff beta}, illustrate that our method exhibits 
insensitivity to the parameter $\lambda$. Across $\lambda \in \{0.05, 0.1, 0.25, 0.5, 1\}$, our method consistently achieves approximately 75\% accuracy on CIFAR10 and 65.5\% on CIFAR100.
This consistent performance highlights our method's ability to deliver stable results regardless of variations in $\lambda$ values, underscoring its robustness to hyperparameter changes.

\textbf{Effectiveness of different objectives}
Our approach comprises two key objectives: vacant-classes distillation and logit suppression.  
The results, presented in Table~\ref{tab: diff objective}, reveal that both vacant-classes distillation and logit suppression contribute to notable performance improvements compared to FedLC~\cite{zhang2022federated}. 
These results demonstrate the effectiveness of our two key objectives in enhancing the overall model performance in federated learning scenarios with significant label skews.

\begin{table}
\caption{Results of combining FedVLS with other methods under Dirichlet-based label skews ($\beta=0.05$) across various datasets. (The values) represent the performance gains.}
\label{table: combination}
\centering
\resizebox{0.48\textwidth}{!}{ 
\begin{tabular}{llll}
\toprule
Method(venue) & \textbf{CIFAR10} & \textbf{CIFAR100} & \textbf{TinyImageNet} \\ \midrule
FedLC (ICML 2022)  & 57.31 & 59.18  & 35.56 \\ 
\rowcolor[gray]{0.9}
+ FedVLS (Ours) & 75.71({\color{red}+18.40}) & 65.71({\color{red}+6.53}) & 45.23({\color{red}+9.67}) \\ 
FedEXP (ICLR 2023) & 63.22 & 62.24 & 34.95 \\ 
\rowcolor[gray]{0.9}
+ FedVLS (Ours) & 75.80({\color{red}+12.58}) & 65.94({\color{red}+3.70}) & 44.76({\color{red}+9.81}) \\ 
FedSAM (ICML2022) & 63.11 & 61.32 & 41.14 \\ 
\rowcolor[gray]{0.9}
+ FedVLS (Ours) & 75.92({\color{red}+12.81}) & 65.46({\color{red}+4.14}) & 48.12({\color{red}+6.98}) \\ 
\bottomrule
\end{tabular}
}
\end{table}

\textbf{Combination with other techniques}
In this section, we integrate our method with two SOTA methods, FedEXP~\cite{jhunjhunwala2023fedexp} and FedSAM~\cite{qu2022generalized}, as detailed in Table~\ref{table: combination}. 
The combination of our method with FedEXP~\cite{jhunjhunwala2023fedexp} and FedSAM~\cite{qu2022generalized} results in improved performance.
This enhancement is reasonable because FedEXP focuses on optimizing the server update for an improved learning rate, and FedSAM emphasizes local gradient descent to achieve a smoother loss landscape. 
These elements complement well with our core idea, which finally results in enhanced performance when combined.

\section{Conclusion}
We have observed that existing federated learning methods always perform poorly in vacant and minority classes, under skewed label distribution across clients.
To overcome these challenges, we introduce FedVLS—an innovative methodology integrating vacant-class distillation and logit suppression simultaneously.
The vacant-class distillation extracts pertinent knowledge regarding vacant classes from the global model for each client, while logit suppression is implemented to directly regularize non-label class logits, addressing the imbalance among majority and minority classes.
Extensive results affirm the effectiveness of both components, surpassing previous state-of-the-art methods across diverse datasets and varying degrees of label skews.
In future work, we will conduct a theoretical analysis of FedVLS, including convergence, privacy, fairness, and other pertinent considerations.

\section{Acknowledgments}
This work was funded by the National Natural Science Foundation of China under Grants (62276256, U2441251) and the Young Elite Scientists Sponsorship Program by CAST (2023QNRC001).

\bibliography{aaai25.bbl}

\begin{thebibliography}{66}
\providecommand{\natexlab}[1]{#1}

\bibitem[{Acar et~al.(2021)Acar, Zhao, Navarro, Mattina, Whatmough, and Saligrama}]{acar2021federated}
Acar, D. A.~E.; Zhao, Y.; Navarro, R.~M.; Mattina, M.; Whatmough, P.~N.; and Saligrama, V. 2021.
\newblock Federated learning based on dynamic regularization.
\newblock \emph{arXiv preprint arXiv:2111.04263}.

\bibitem[{Chai et~al.(2023)Chai, Wang, Yang, Zhang, Chen, and Yang}]{chai2023survey}
Chai, D.; Wang, L.; Yang, L.; Zhang, J.; Chen, K.; and Yang, Q. 2023.
\newblock A Survey for Federated Learning Evaluations: Goals and Measures.
\newblock \emph{arXiv preprint arXiv:2308.11841}.

\bibitem[{Chen et~al.(2022)Chen, Liu, Ma, and Lyu}]{chen2022calfat}
Chen, C.; Liu, Y.; Ma, X.; and Lyu, L. 2022.
\newblock Calfat: Calibrated federated adversarial training with label skewness.
\newblock In \emph{Proc. NeurIPS}.

\bibitem[{Chen, Vikalo et~al.(2023)}]{chen2023best}
Chen, H.; Vikalo, H.; et~al. 2023.
\newblock The Best of Both Worlds: Accurate Global and Personalized Models through Federated Learning with Data-Free Hyper-Knowledge Distillation.
\newblock In \emph{Proc. ICLR}.

\bibitem[{Cubuk et~al.(2019)Cubuk, Zoph, Mane, Vasudevan, and Le}]{cubuk2018autoaugment}
Cubuk, E.~D.; Zoph, B.; Mane, D.; Vasudevan, V.; and Le, Q.~V. 2019.
\newblock Autoaugment: Learning augmentation policies from data.
\newblock In \emph{Proc. CVPR}.

\bibitem[{Cui et~al.(2019)Cui, Jia, Lin, Song, and Belongie}]{cui2019class}
Cui, Y.; Jia, M.; Lin, T.-Y.; Song, Y.; and Belongie, S. 2019.
\newblock Class-balanced loss based on effective number of samples.
\newblock In \emph{Proc. CVPR}.

\bibitem[{Deng(2012)}]{deng2012mnist}
Deng, L. 2012.
\newblock The mnist database of handwritten digit images for machine learning research.
\newblock \emph{Proc. SPM}.

\bibitem[{Diao, Li, and He(2024)}]{diao2024exploiting}
Diao, Y.; Li, Q.; and He, B. 2024.
\newblock Exploiting Label Skews in Federated Learning with Model Concatenation.
\newblock In \emph{Proc. AAAI}.

\bibitem[{Divyansh~Jhunjhunwala(2023)}]{jhunjhunwala2023fedexp}
Divyansh~Jhunjhunwala, G.~J., Shiqiang~Wang. 2023.
\newblock FedExP: Speeding up Federated Averaging Via Extrapolation.
\newblock In \emph{Proc. ICLR}.

\bibitem[{Fan et~al.(2024)Fan, Yao, Han, Zhang, Wang et~al.}]{fan2024federated}
Fan, Z.; Yao, J.; Han, B.; Zhang, Y.; Wang, Y.; et~al. 2024.
\newblock Federated Learning with Bilateral Curation for Partially Class-Disjoint Data.
\newblock \emph{Proc. NeurIPS}.

\bibitem[{Fan et~al.(2023)Fan, Yao, Zhang, Lyu, Wang, and Zhang}]{fan2023federated}
Fan, Z.; Yao, J.; Zhang, R.; Lyu, L.; Wang, Y.; and Zhang, Y. 2023.
\newblock Federated Learning under Partially Disjoint Data via Manifold Reshaping.
\newblock \emph{Proc. JMLR}.

\bibitem[{Guo, Wang, and Geng(2024)}]{guo2024dynamic}
Guo, S.; Wang, H.; and Geng, X. 2024.
\newblock Dynamic heterogeneous federated learning with multi-level prototypes.
\newblock \emph{Proc. PR}.

\bibitem[{Guo et~al.(2024)Guo, Wang, Lin, Kou, and Geng}]{guo2024addressing}
Guo, S.; Wang, H.; Lin, S.; Kou, Z.; and Geng, X. 2024.
\newblock Addressing Skewed Heterogeneity via Federated Prototype Rectification With Personalization.
\newblock \emph{Proc. TNNLS}.

\bibitem[{Hsu, Qi, and Brown(2019)}]{hsu2019measuring}
Hsu, T.-M.~H.; Qi, H.; and Brown, M. 2019.
\newblock Measuring the effects of non-identical data distribution for federated visual classification.
\newblock \emph{arXiv preprint arXiv:1909.06335}.

\bibitem[{Huang et~al.(2023)Huang, Ye, Shi, Li, and Du}]{huang2023rethinking}
Huang, W.; Ye, M.; Shi, Z.; Li, H.; and Du, B. 2023.
\newblock Rethinking federated learning with domain shift: A prototype view.
\newblock In \emph{Proc. CVPR}.

\bibitem[{Itahara et~al.(2021)Itahara, Nishio, Koda, Morikura, and Yamamoto}]{itahara2021distillation}
Itahara, S.; Nishio, T.; Koda, Y.; Morikura, M.; and Yamamoto, K. 2021.
\newblock Distillation-based semi-supervised federated learning for communication-efficient collaborative training with non-iid private data.
\newblock \emph{IEEE Transactions on Mobile Computing}.

\bibitem[{Jeong et~al.(2018)Jeong, Oh, Kim, Park, Bennis, and Kim}]{jeong2018communication}
Jeong, E.; Oh, S.; Kim, H.; Park, J.; Bennis, M.; and Kim, S.-L. 2018.
\newblock Communication-efficient on-device machine learning: Federated distillation and augmentation under non-iid private data.
\newblock In \emph{Proc. NeurIPS Workshops}.

\bibitem[{Kairouz et~al.(2021)Kairouz, McMahan, Avent, Bellet, Bennis, Bhagoji, Bonawitz, Charles, Cormode, Cummings et~al.}]{kairouz2021advances}
Kairouz, P.; McMahan, H.~B.; Avent, B.; Bellet, A.; Bennis, M.; Bhagoji, A.~N.; Bonawitz, K.; Charles, Z.; Cormode, G.; Cummings, R.; et~al. 2021.
\newblock Advances and open problems in federated learning.
\newblock \emph{Foundations and Trends{\textregistered} in Machine Learning}.

\bibitem[{Karimireddy et~al.(2020)Karimireddy, Kale, Mohri, Reddi, Stich, and Suresh}]{karimireddy2020scaffold}
Karimireddy, S.~P.; Kale, S.; Mohri, M.; Reddi, S.; Stich, S.; and Suresh, A.~T. 2020.
\newblock Scaffold: Stochastic controlled averaging for federated learning.
\newblock In \emph{Proc. ICML}.

\bibitem[{Kone{\v{c}}n{\`y} et~al.(2016)Kone{\v{c}}n{\`y}, McMahan, Ramage, and Richt{\'a}rik}]{konevcny2016federated}
Kone{\v{c}}n{\`y}, J.; McMahan, H.~B.; Ramage, D.; and Richt{\'a}rik, P. 2016.
\newblock Federated optimization: Distributed machine learning for on-device intelligence.
\newblock \emph{arXiv preprint arXiv:1610.02527}.

\bibitem[{Krizhevsky(2009)}]{krizhevsky2009learning}
Krizhevsky, A. 2009.
\newblock Learning Multiple Layers of Features from Tiny Images.
\newblock \emph{Master's thesis, University of Tront}.

\bibitem[{Le and Yang(2015)}]{le2015tiny}
Le, Y.; and Yang, X. 2015.
\newblock Tiny imagenet visual recognition challenge.
\newblock \emph{CS 231N}.

\bibitem[{Lee et~al.(2022)Lee, Jeong, Shin, Bae, and Yun}]{lee2022preservation}
Lee, G.; Jeong, M.; Shin, Y.; Bae, S.; and Yun, S.-Y. 2022.
\newblock Preservation of the global knowledge by not-true distillation in federated learning.
\newblock In \emph{Proc. NeurIPS}.

\bibitem[{Lee and Yoon(2024)}]{leerethinking}
Lee, T.; and Yoon, S.~W. 2024.
\newblock Rethinking the Flat Minima Searching in Federated Learning.
\newblock In \emph{Proc. ICML}.

\bibitem[{Li et~al.(2023)Li, Schmidt, Alstr{\o}m, and Stich}]{li2023effectiveness}
Li, B.; Schmidt, M.~N.; Alstr{\o}m, T.~S.; and Stich, S.~U. 2023.
\newblock On the effectiveness of partial variance reduction in federated learning with heterogeneous data.
\newblock In \emph{Proc. CVPR}.

\bibitem[{Li and Wang(2019)}]{li2019fedmd}
Li, D.; and Wang, J. 2019.
\newblock Fedmd: Heterogenous federated learning via model distillation.
\newblock In \emph{Proc. NeurIPS Workshops}.

\bibitem[{Li et~al.(2022)Li, Diao, Chen, and He}]{li2022federated}
Li, Q.; Diao, Y.; Chen, Q.; and He, B. 2022.
\newblock Federated learning on non-iid data silos: An experimental study.
\newblock In \emph{Proc. ICDE}.

\bibitem[{Li, He, and Song(2021)}]{li2021model}
Li, Q.; He, B.; and Song, D. 2021.
\newblock Model-contrastive federated learning.
\newblock In \emph{Proc. CVPR}.

\bibitem[{Li et~al.(2020{\natexlab{a}})Li, Sahu, Talwalkar, and Smith}]{li2020federated1}
Li, T.; Sahu, A.~K.; Talwalkar, A.; and Smith, V. 2020{\natexlab{a}}.
\newblock Federated learning: Challenges, methods, and future directions.
\newblock \emph{Proc. SPM}.

\bibitem[{Li et~al.(2020{\natexlab{b}})Li, Sahu, Zaheer, Sanjabi, Talwalkar, and Smith}]{li2020federated}
Li, T.; Sahu, A.~K.; Zaheer, M.; Sanjabi, M.; Talwalkar, A.; and Smith, V. 2020{\natexlab{b}}.
\newblock Federated optimization in heterogeneous networks.
\newblock In \emph{Proc. MLSys}.

\bibitem[{Li et~al.(2019)Li, Huang, Yang, Wang, and Zhang}]{li2019convergence}
Li, X.; Huang, K.; Yang, W.; Wang, S.; and Zhang, Z. 2019.
\newblock On the convergence of fedavg on non-iid data.
\newblock In \emph{Proc. ICLR}.

\bibitem[{Li and Zhan(2021)}]{li2021fedrs}
Li, X.-C.; and Zhan, D.-C. 2021.
\newblock Fedrs: Federated learning with restricted softmax for label distribution non-iid data.
\newblock In \emph{Proc. KDD}.

\bibitem[{Lin et~al.(2020)Lin, Kong, Stich, and Jaggi}]{lin2020ensemble}
Lin, T.; Kong, L.; Stich, S.~U.; and Jaggi, M. 2020.
\newblock Ensemble distillation for robust model fusion in federated learning.
\newblock In \emph{Proc. NeurIPS}.

\bibitem[{Liu et~al.(2022)Liu, Bai, Yu, and Zhang}]{liu2022towards}
Liu, D.; Bai, L.; Yu, T.; and Zhang, A. 2022.
\newblock Towards Method of Horizontal Federated Learning: A Survey.
\newblock In \emph{Proc. BigDIA}.

\bibitem[{Lu et~al.(2023)Lu, Li, Bao, Wang, Qian, and Ge}]{lu2023federated}
Lu, J.; Li, S.; Bao, K.; Wang, P.; Qian, Z.; and Ge, S. 2023.
\newblock Federated Learning with Label-Masking Distillation.
\newblock In \emph{Proc. ACM-MM}.

\bibitem[{Luo et~al.(2023)Luo, Wang, Fu, Li, Lan, and Gao}]{luo2023dfrd}
Luo, K.; Wang, S.; Fu, Y.; Li, X.; Lan, Y.; and Gao, M. 2023.
\newblock DFRD: Data-Free Robustness Distillation for Heterogeneous Federated Learning.
\newblock In \emph{Proc. NeurIPS}.

\bibitem[{Luo et~al.(2021)Luo, Chen, Hu, Zhang, Liang, and Feng}]{luo2021no}
Luo, M.; Chen, F.; Hu, D.; Zhang, Y.; Liang, J.; and Feng, J. 2021.
\newblock No fear of heterogeneity: Classifier calibration for federated learning with non-iid data.
\newblock In \emph{Proc. NeurIPS}.

\bibitem[{Luo, Wang, and Wang(2024)}]{luo2024federated}
Luo, Z.; Wang, Y.; and Wang, Z. 2024.
\newblock Federated Local Compact Representation Communication: Framework and Application.
\newblock \emph{Proc. MIR}.

\bibitem[{Luo et~al.(2022)Luo, Wang, Wang, Sun, and Tan}]{luo2022disentangled}
Luo, Z.; Wang, Y.; Wang, Z.; Sun, Z.; and Tan, T. 2022.
\newblock Disentangled federated learning for tackling attributes skew via invariant aggregation and diversity transferring.
\newblock \emph{arXiv preprint arXiv:2206.06818}.

\bibitem[{Ma et~al.(2023)Ma, Jiao, Liu, Yang, Liu, and Li}]{Ma_2023_CVPR}
Ma, Y.; Jiao, L.; Liu, F.; Yang, S.; Liu, X.; and Li, L. 2023.
\newblock Curvature-Balanced Feature Manifold Learning for Long-Tailed Classification.
\newblock In \emph{Proc. CVPR}.

\bibitem[{McMahan et~al.(2017)McMahan, Moore, Ramage, Hampson, and y~Arcas}]{mcmahan2017communication}
McMahan, B.; Moore, E.; Ramage, D.; Hampson, S.; and y~Arcas, B.~A. 2017.
\newblock Communication-efficient learning of deep networks from decentralized data.
\newblock In \emph{Proc. AISTATS}.

\bibitem[{Menon et~al.(2021)Menon, Jayasumana, Rawat, Jain, Veit, and Kumar}]{menon2020long}
Menon, A.~K.; Jayasumana, S.; Rawat, A.~S.; Jain, H.; Veit, A.; and Kumar, S. 2021.
\newblock Long-tail learning via logit adjustment.
\newblock In \emph{Proc. ICLR}.

\bibitem[{Mu et~al.(2023)Mu, Shen, Cheng, Geng, Fu, Zhang, and Zhang}]{mu2023fedproc}
Mu, X.; Shen, Y.; Cheng, K.; Geng, X.; Fu, J.; Zhang, T.; and Zhang, Z. 2023.
\newblock Fedproc: Prototypical contrastive federated learning on non-iid data.
\newblock \emph{Proc. FGCS}.

\bibitem[{Qu et~al.(2022)Qu, Li, Duan, Liu, Tang, and Lu}]{qu2022generalized}
Qu, Z.; Li, X.; Duan, R.; Liu, Y.; Tang, B.; and Lu, Z. 2022.
\newblock Generalized federated learning via sharpness aware minimization.
\newblock In \emph{Proc. ICML}.

\bibitem[{Reguieg et~al.(2023)Reguieg, El~Hanjri, El~Kamili, and Kobbane}]{reguieg2023comparative}
Reguieg, H.; El~Hanjri, M.; El~Kamili, M.; and Kobbane, A. 2023.
\newblock A Comparative Evaluation of FedAvg and Per-FedAvg Algorithms for Dirichlet Distributed Heterogeneous Data.
\newblock In \emph{Proc. WINCOM}.

\bibitem[{Sandler et~al.(2018)Sandler, Howard, Zhu, Zhmoginov, and Chen}]{sandler2018mobilenetv2}
Sandler, M.; Howard, A.; Zhu, M.; Zhmoginov, A.; and Chen, L.-C. 2018.
\newblock Mobilenetv2: Inverted residuals and linear bottlenecks.
\newblock In \emph{Proc. CVPR}.

\bibitem[{Sheller et~al.(2020)Sheller, Edwards, Reina, Martin, Pati, Kotrotsou, Milchenko, Xu, Marcus, Colen et~al.}]{sheller2020federated}
Sheller, M.~J.; Edwards, B.; Reina, G.~A.; Martin, J.; Pati, S.; Kotrotsou, A.; Milchenko, M.; Xu, W.; Marcus, D.; Colen, R.~R.; et~al. 2020.
\newblock Federated learning in medicine: facilitating multi-institutional collaborations without sharing patient data.
\newblock \emph{Scientific reports}.

\bibitem[{Shen, Wang, and Lv(2023)}]{shen2023federated}
Shen, Y.; Wang, H.; and Lv, H. 2023.
\newblock Federated Learning with Classifier Shift for Class Imbalance.
\newblock \emph{arXiv preprint arXiv:2304.04972}.

\bibitem[{Shi et~al.(2023{\natexlab{a}})Shi, Liang, Zhang, Tan, and Bai}]{shi2022towards}
Shi, Y.; Liang, J.; Zhang, W.; Tan, V.~Y.; and Bai, S. 2023{\natexlab{a}}.
\newblock Towards Understanding and Mitigating Dimensional Collapse in Heterogeneous Federated Learning.
\newblock In \emph{Proc. ICLR}.

\bibitem[{Shi et~al.(2023{\natexlab{b}})Shi, Liang, Zhang, Xue, Tan, and Bai}]{shi2023understanding}
Shi, Y.; Liang, J.; Zhang, W.; Xue, C.; Tan, V.~Y.; and Bai, S. 2023{\natexlab{b}}.
\newblock Understanding and Mitigating Dimensional Collapse in Federated Learning.
\newblock \emph{IEEE Transactions on Pattern Analysis and Machine Intelligence}.

\bibitem[{T~Dinh, Tran, and Nguyen(2020)}]{t2020personalized}
T~Dinh, C.; Tran, N.; and Nguyen, J. 2020.
\newblock Personalized federated learning with moreau envelopes.
\newblock In \emph{Proc. NeurIPS}.

\bibitem[{Tan et~al.(2020)Tan, Wang, Li, Li, Ouyang, Yin, and Yan}]{tan2020equalization}
Tan, J.; Wang, C.; Li, B.; Li, Q.; Ouyang, W.; Yin, C.; and Yan, J. 2020.
\newblock Equalization loss for long-tailed object recognition.
\newblock In \emph{Proc. CVPR}.

\bibitem[{Wang et~al.(2023{\natexlab{a}})Wang, Li, Xu, Li, Zhan, and Zeng}]{wang2023dafkd}
Wang, H.; Li, Y.; Xu, W.; Li, R.; Zhan, Y.; and Zeng, Z. 2023{\natexlab{a}}.
\newblock DaFKD: Domain-aware Federated Knowledge Distillation.
\newblock In \emph{Proc. CVPR}.

\bibitem[{Wang et~al.(2023{\natexlab{b}})Wang, Li, Tan, Jiang, Sun, Liu, Gao, and Wu}]{wang2023federated}
Wang, Y.; Li, R.; Tan, H.; Jiang, X.; Sun, S.; Liu, M.; Gao, B.; and Wu, Z. 2023{\natexlab{b}}.
\newblock Federated Skewed Label Learning with Logits Fusion.
\newblock \emph{arXiv preprint arXiv:2311.08202}.

\bibitem[{Weyand et~al.(2020)Weyand, Araujo, Cao, and Sim}]{weyand2020google}
Weyand, T.; Araujo, A.; Cao, B.; and Sim, J. 2020.
\newblock Google landmarks dataset v2-a large-scale benchmark for instance-level recognition and retrieval.
\newblock In \emph{Proc. CVPR}.

\bibitem[{Wu et~al.(2023)Wu, Sun, Wang, Liu, Jiang, and Li}]{wu2023survey}
Wu, Z.; Sun, S.; Wang, Y.; Liu, M.; Jiang, X.; and Li, R. 2023.
\newblock Survey of Knowledge Distillation in Federated Edge Learning.
\newblock \emph{arXiv preprint arXiv:2301.05849}.

\bibitem[{Xiao et~al.(2023)Xiao, Chen, Liu, Wang, Feng, Hao, Zhou, Wu, Yang, and Liu}]{xiao2023fed}
Xiao, Z.; Chen, Z.; Liu, S.; Wang, H.; Feng, Y.; Hao, J.; Zhou, J.~T.; Wu, J.; Yang, H.~H.; and Liu, Z. 2023.
\newblock Fed-GraB: Federated Long-tailed Learning with Self-Adjusting Gradient Balancer.
\newblock \emph{arXiv preprint arXiv:2310.07587}.

\bibitem[{Yang, Fang, and Liu(2021)}]{yang2021achieving}
Yang, H.; Fang, M.; and Liu, J. 2021.
\newblock Achieving linear speedup with partial worker participation in non-iid federated learning.
\newblock In \emph{Proc. ICLR}.

\bibitem[{Ye et~al.(2023)Ye, Fang, Du, Yuen, and Tao}]{ye2023heterogeneous}
Ye, M.; Fang, X.; Du, B.; Yuen, P.~C.; and Tao, D. 2023.
\newblock Heterogeneous federated learning: State-of-the-art and research challenges.
\newblock \emph{ACM Computing Surveys}.

\bibitem[{Yeganeh et~al.(2020)Yeganeh, Farshad, Navab, and Albarqouni}]{yeganeh2020inverse}
Yeganeh, Y.; Farshad, A.; Navab, N.; and Albarqouni, S. 2020.
\newblock Inverse distance aggregation for federated learning with non-iid data.
\newblock In \emph{Proc. MICCAI Workshops}.

\bibitem[{Zeng et~al.(2023)Zeng, Liu, Liu, Shen, Liu, and Wu}]{zeng2023global}
Zeng, Y.; Liu, L.; Liu, L.; Shen, L.; Liu, S.; and Wu, B. 2023.
\newblock Global Balanced Experts for Federated Long-Tailed Learning.
\newblock In \emph{Proc. ICCV}.

\bibitem[{Zhang et~al.(2023)Zhang, Li, Qi, and He}]{zhang2023survey}
Zhang, J.; Li, C.; Qi, J.; and He, J. 2023.
\newblock A Survey on Class Imbalance in Federated Learning.
\newblock \emph{arXiv preprint arXiv:2303.11673}.

\bibitem[{Zhang et~al.(2022)Zhang, Li, Li, Xu, Wu, Ding, and Wu}]{zhang2022federated}
Zhang, J.; Li, Z.; Li, B.; Xu, J.; Wu, S.; Ding, S.; and Wu, C. 2022.
\newblock Federated learning with label distribution skew via logits calibration.
\newblock In \emph{Proc. ICML}.

\bibitem[{Zhao et~al.(2022)Zhao, Cui, Song, Qiu, and Liang}]{zhao2022decoupled}
Zhao, B.; Cui, Q.; Song, R.; Qiu, Y.; and Liang, J. 2022.
\newblock Decoupled knowledge distillation.
\newblock In \emph{Proc. CVPR}.

\bibitem[{Zhu et~al.(2021)Zhu, Xu, Liu, and Jin}]{zhu2021federated}
Zhu, H.; Xu, J.; Liu, S.; and Jin, Y. 2021.
\newblock Federated learning on non-IID data: A survey.
\newblock \emph{Neurocomputing}.

\bibitem[{Zhu, Hong, and Zhou(2021)}]{zhu2021data}
Zhu, Z.; Hong, J.; and Zhou, J. 2021.
\newblock Data-free knowledge distillation for heterogeneous federated learning.
\newblock In \emph{Proc. ICML}.

\end{thebibliography}

\newpage
\twocolumn[
\begin{@twocolumnfalse}
\section*{\centering {Technical Appendix} \vspace{20pt}}
\end{@twocolumnfalse}
]

\section{The Pseudocode of Our Method}
\label{appendixB}
\begin{algorithm}
    \renewcommand{\algorithmicrequire}{\textbf{Input:}}
    \renewcommand{\algorithmicensure}{\textbf{Output:}}
    \caption{\textbf{FedVLS}}
    \label{alg:FedVLS}
    \begin{algorithmic}[1]
        \REQUIRE number of communication rounds $T$, number of clients $N$, client participating rate $R$, number of local epochs $E$, batch size $B$, learning rate $\eta$.
        \ENSURE the global model $\boldsymbol{\mathcal{\omega}}^T$
        \STATE initialize $\boldsymbol{\mathcal{\omega}}^0$
        \STATE $m \gets \max( \lfloor R \cdot N \rfloor, 1)$
        \FOR{\textit{communication round} $t=1,2,\cdots,T-1$}
            \STATE $M_t \gets $ randomly select a subset containing $m$ clients   
            \FOR{\textit{each client} $i \in M_t$}
                \STATE $\boldsymbol{\mathcal{\omega}}^t_i =  \boldsymbol{\mathcal{\omega}}^t$
                \STATE $\boldsymbol{\mathcal{\omega}}^{t+1}_i \gets$ \textbf{LocalUpdate}($\boldsymbol{\mathcal{\omega}}^t_i$)
            \ENDFOR
            \STATE $\boldsymbol{\mathcal{\omega}}^{t+1} = \boldsymbol{\mathcal{\omega}}^t + \sum_{i \in M_t}{\frac{|\mathcal{D}_i|}{|\mathcal{D}|}(\boldsymbol{\mathcal{\omega}}^{t+1}_i - \boldsymbol{\mathcal{\omega}}^t_i)}$
        \ENDFOR 
        \vspace{0.5cm}
        \STATE \textbf{LocalUpdate}($\boldsymbol{\mathcal{\omega}}^t_i$):
        \FOR{\textit{epoch} $e=1,2,\cdots,E$}
            \FOR{\textit{each batch} $\mathcal{B}_{i} = \{\textit{x},\textit{y}\} \in \mathcal{D}_i$} \vspace{0.4\baselineskip}
                \STATE $\mathcal{L}_{\bf cal}(\bf\boldsymbol{\omega};\mathcal{B}_\textit{i}) = -\mathbb{E}_{(\textit{x},\textit{y}) \sim \mathcal{B}_\textit{i}} \log \left(\frac{\textit{p}(\textit{y}) \cdot e^{\textit{f}(\textit{x};{\bf\boldsymbol{\omega}})\left[\textit{y}\right]}}{\sum_{\textit{c}}{\textit{p}(\textit{c}) \cdot e^{\textit{f}(\textit{x};{\bf\boldsymbol{\omega}})[\textit{c}]}}}\right)$ \vspace{0.4\baselineskip}
                \STATE $\mathcal{L}_{\bf dis}(\bf\boldsymbol{\omega};\mathcal{B}_\textit{i}) = \mathbb{E}_{(\textit{x},\textit{y}) \sim \mathcal{B}_\textit{i}} \sum_{\textit{o} \in \mathbb{O}} {\textit{q}^g(\textit{o};\textit{x}) \log \left[ \frac{\textit{q}(\textit{o};\textit{x})}{\textit{q}^g(\textit{o};\textit{x})} \right]}$ \vspace{0.4\baselineskip} 
                \STATE $\mathcal{L}^{c}_{\bf logit}(\bf\boldsymbol{\omega};\mathcal{B}_\textit{i}) = \log \left(\mathbb{E}_{(\textit{x},\textit{y}) \sim \mathcal{B}_\textit{i}} \mathbb{I}({\textit{y} \neq \textit{c}}) \cdot e^{\textit{f}(\textit{x};\boldsymbol{\omega}) [\textit{c}]}\right)$\vspace{0.4\baselineskip}
                \STATE $\mathcal{L}_{\bf logit}(\bf\boldsymbol{\omega};\mathcal{B}_\textit{i}) = \sum \textit{p}(\textit{c}) \cdot \mathcal{L}^{\textit{c}}_{\bf logit}(\bf\boldsymbol{\omega};\mathcal{B}_\textit{i})$\vspace{0.4\baselineskip}
                \STATE $\mathcal{L}({\bf \boldsymbol{\omega}}^t_i;\mathcal{B}_i) = \mathcal{L}_{\bf cal}({\bf \boldsymbol{\omega}}^t_i;\mathcal{B}_i) + \lambda \cdot \mathcal{L}_{\bf dis}({\bf \boldsymbol{\omega}}^t_i;\mathcal{B}_i) + \mathcal{L}_{\bf logit}({\bf \boldsymbol{\omega}}^t_i;\mathcal{B}_i)$ \vspace{0.4\baselineskip}
                \STATE $\boldsymbol{\mathcal{\omega}}^t_i = \boldsymbol{\mathcal{\omega}}^t_i - \eta \nabla {\mathcal{L}({\bf \boldsymbol{\omega}}^t_i;\mathcal{B}_i)}$
            \ENDFOR
        \ENDFOR 
        \STATE \textbf{return} $\boldsymbol{\mathcal{\omega}}^t_i$
\end{algorithmic}
\end{algorithm}

\section{Experimental Details}
\begin{figure*}[ht]
  \centering
  \includegraphics[width=0.95\textwidth]{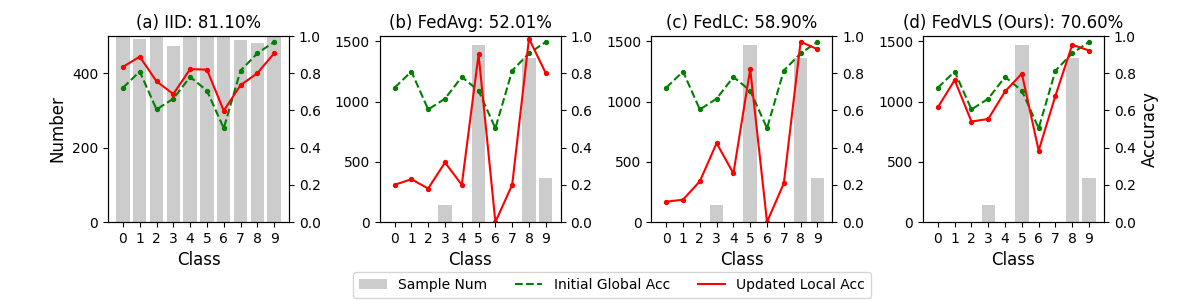}
  \caption{Class-wise accuracy of the initial global model and updated local model on IID and label-skewed CIFAR10 data distributions. (a) represents the result updating on IID local data with FedAvg~\cite{mcmahan2017communication}. (b-d) showcase the results updating on skewed local data distribution with FedAvg, FedLC~\cite{zhang2022federated}, and our FedVLS, respectively. The value (\%) in each caption corresponds to the accuracy of the global model aggregated from updated local models.} 
  \label{fig: class-wise acc on client0 s}
\end{figure*}

\subsection{Data Distribution among Clients}
In Figure 1 (a) of the main paper, all clients' data distributions are independent and identically sampled. In Figures\ref{fig: class-wise acc on client0} (b), (c), (d) of the main paper, the data distribution of all clients is shown in Table ~\ref{tab: data distribution} as follows. We focus on client 0 for analysis, where it is evident that classes 5, 8, and 9 are majority classes, class 3 is a minority class, and the remaining classes are vacant.

\begin{figure}[ht]
  \centering
  \includegraphics[width=0.48\textwidth]{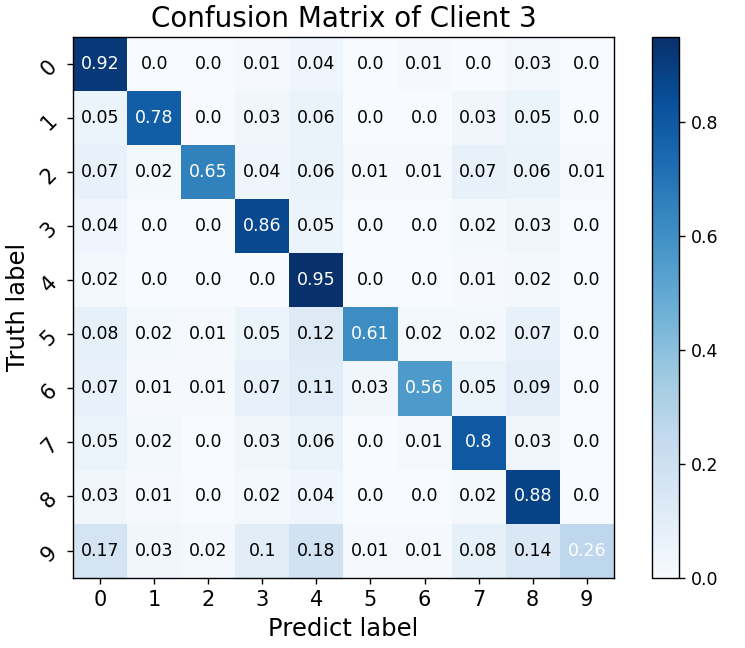}
  \caption{Confusion matrix of client 3 on CIFAR10 dataset with Dirichlet-based label skews ($\beta$ = 0.5) using FedLC~\cite{zhang2022federated}.}
  \label{fig: confusion s}
\end{figure}

\begin{figure*}[h!]
    \centering
    \begin{minipage}{0.48\textwidth}
        \centering
        \begin{subfigure}[b]{0.95\textwidth}
            \includegraphics[width=\textwidth]{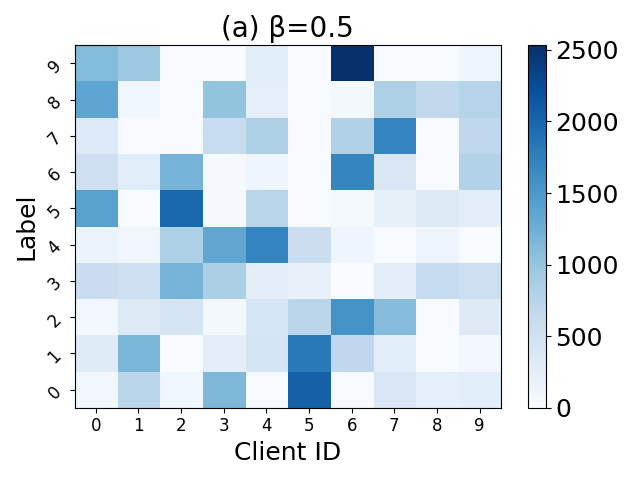}
        \end{subfigure}
        \par\medskip
        \begin{subfigure}[b]{0.95\textwidth}
            \includegraphics[width=\textwidth]{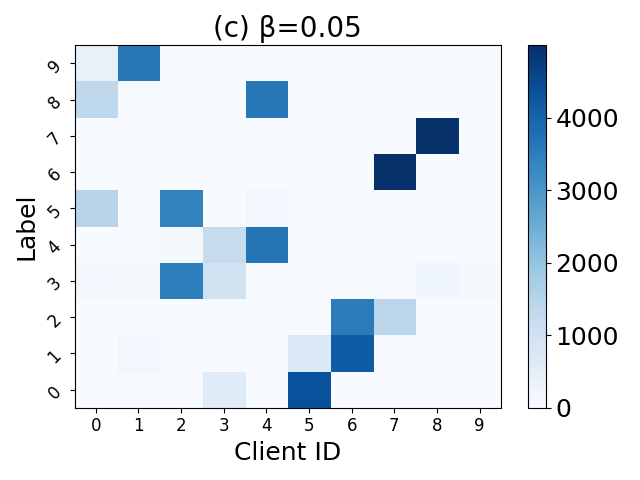}
        \end{subfigure}
    \end{minipage}\hfill
    \begin{minipage}{0.48\textwidth}
        \centering
        \begin{subfigure}[b]{0.95\textwidth}
            \includegraphics[width=\textwidth]{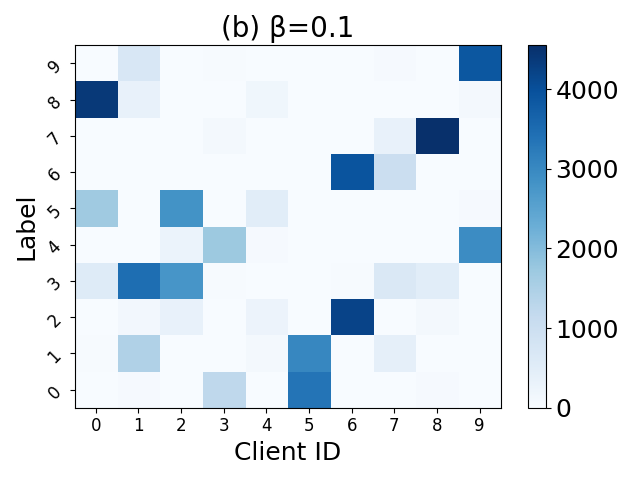}
        \end{subfigure}
        \par\medskip
        \begin{subfigure}[b]{0.95\textwidth}
            \includegraphics[width=\textwidth]{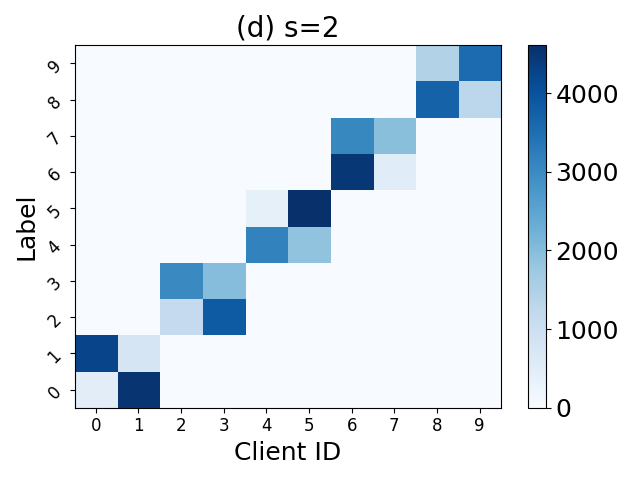}
        \end{subfigure}
    \end{minipage}
    \caption{Visualization of the Dirichlet-based ($\beta = 0.5, 0.1, 0.05$) and quantity-based (s=2) label skews of CIFAR10 dataset among 10 clients.}
    \label{fig: heat}
\end{figure*}

In Figure 2 of the main paper, the data distribution for this client is shown in the fourth column of Figure~\ref{fig: heat} (a). Here, classes 0, 1, 3, and 7 are majority classes, while classes 2, 5, and 6 are minority classes. Figure~\ref{fig: confusion s} reveals that minority classes are frequently misclassified as majority classes, which motivates the introduction of Logit Suppression in the main paper.

In our experiments, we incorporate Dirichlet-based label skews ($\beta = 0.5, 0.1, 0.05$) and quantity-based label skews (s=2) for the CIFAR10 dataset. The data distribution for these skews is illustrated in Figure~\ref{fig: heat}.

\begin{table}[ht]
    \centering
    \caption{The data distribution among clients with Dirichlet-based ($\beta=0.1$) CIFAR10 datasets.}
    \resizebox{0.48\textwidth}{!}{
    \begin{tabular}{lcccccccccc}
    \toprule
    client &  0 &  1 &  2 & 3 &  4 & 5 &  6 &  7 &  8 & 9  \\  \midrule
    class 0 & 0 & 57 & 0 & 600 & 0 & 4342 & 0 & 0 & 0 & 1  \\ 
    class 1 & 0 & 155 & 0 & 0 & 1 & 679 & 4153 & 0 & 11 & 1  \\ 
    class 2 & 0 & 3 & 24 & 0 & 15 & 0 & 3536 & 1419 & 0 & 3  \\ 
    class 3 & 141 & 99 & 3490 & 953 & 0 & 0 & 0 & 0 & 208 & 109  \\ 
    class 4 & 0 & 0 & 98 & 1217 & 3684 & 0 & 0 & 0 & 1 & 0  \\ 
    class 5 & 1471 & 0 & 3403 & 0 & 125 & 0 & 0 & 0 & 0 & 1  \\ 
    class 6 & 0 & 0 & 0 & 0 & 0 & 0 & 0 & 4999 & 1 & 0  \\ 
    class 7 & 0 & 0 & 0 & 2 & 0 & 0 & 0 & 0 & 4998 & 0  \\ 
    class 8 & 1360 & 35 & 0 & 0 & 3604 & 0 & 0 & 0 & 0 & 1  \\ 
    class 9 & 366 & 4608 & 0 & 0 & 0 & 0 & 0 & 0 & 0 & 26  \\ 
    \bottomrule
    \label{tab: data distribution}
    \end{tabular}
    }
\end{table}

\subsection{Implementation Details}
The augmentation for all CIFAR and TinyImageNet experiments is the same as existing literature AutoAugment~\cite{cubuk2018autoaugment}.
The specific architecture of MobileNetV2~\cite{sandler2018mobilenetv2} is shown in Table~\ref{tab: model construction}, while the structure of the bottleneck is detailed in Table~\ref{tab: bottleneck construction}.
Since the architectures of ResNet-18 and ResNet-32 are well-known, we do not present their detailed structures here.
Hyperparameters for all baseline methods are set according to the configurations specified in the original papers, as detailed in Table~\ref{table: paraemters}. 
All experiments are conducted on a single NVIDIA GeForce RTX 3090 with 24GB of memory.

\begin{table}[ht]
\centering
\caption{The architecture of MobileNetV2.}
\label{tab: model construction}
\begin{tabular}{c|c|c|c|c|c}
\toprule
Input & Operator & t & *C* & *n* & s \\ \midrule
$224^{2}\times 3$ & conv2d & - & 32 & 1 & 2 \\
$112^{2}\times 32$ & bottleneck & 1 & 16 & 1 & 1 \\
$112^{2}\times 16$ & bottleneck & 6 & 24 & 2 & 2 \\
$56^{2}\times 24$ & bottleneck & 6 & 32 & 3 & 2 \\
$28^{2}\times 32$ & bottleneck & 6 & 64 & 4 & 2 \\
$14^{2}\times 64$ & bottleneck & 6 & 96 & 3 & 1 \\
$14^{2}\times 96$ & bottleneck & 6 & 160 & 3 & 2 \\
$7^{2}\times 160$ & bottleneck & 6 & 320 & 1 & 1 \\
$7^{2}\times 320$ & conv$2d1\times 1$ & - & 1280 & 1 & 1 \\
$7^{2}\times 1280$ & avgpool$7\times 7$ & - & - & 1 & - \\
$1\times 1\times 1280$ & conv$2d\space 1\times 1$ & - & k & - & \\ \bottomrule
\end{tabular}
\end{table}

\begin{table}[h]
\centering
\caption{The architecture of bottleneck.}
\label{tab: bottleneck construction}
\resizebox{0.48\textwidth}{!}{
\begin{tabular}{c|c|c}
\toprule
Input & Operator & Output \\ \midrule
$h \times w \times k$ & 1$\times$1 conv2d, ReLU6 & $h \times w \times (tk)$ \\n$h \times w \times tk$ & $3 \times 3$ dwise s=s, ReLU6 & $\frac{h}{s} \times \frac{w}{s} \times (tk)$ $\frac{h}{s} \times \frac{w}{s} \times k^{\prime}$ \\n$\frac{h}{s} \times \frac{w}{s} \times tk$ & linear$1 \times 1$ conv2d & $\frac{h}{s} \times \frac{w}{s} \times (tk)$ $\frac{h}{s} \times \frac{w}{s} \times k^{\prime}$ \\ \bottomrule
\end{tabular}
}
\end{table}

\begin{table}[ht]
\centering
\caption{The hyperparameters for all baseline methods.}
\resizebox{0.35\textwidth}{!}{ 
\begin{tabular}{lc}
\toprule
FedAvg (AISTATS 2017) &  None \\
FedProx (MLSys 2020) &  $\mu$=0.01 \\
MOON (CVPR 2021) &  $\mu$=0.01, $\tau$=0.5 \\
FedEXP (ICLR 2023) &  $\epsilon$=0.01\\
FedLC (ICML 2022) &  $\tau$=0.5 \\
FedRS (KDD 2021) &  $\alpha$=0.7 \\
FedSAM (ICML2022) & $\rho$=0.1, $\beta$=0.9 \\
FedNTD (NeurIPS 2022) &  $\beta$=0.1\\ 
FedMR (TMLR 2023) &  $deco$=4\\ 
FedLMD (MM 2023) &  $\beta$=0.1\\
FedConcat (AAAI 2024) &  $cluster$=\{2, 4\}\\
FedGF (ICML 2024) &  $\rho$=0.1, $c_os$=0.3 \\ \bottomrule
\end{tabular}
}
\label{table: paraemters}
\end{table}

\section{Additional Experimental Observations}
In Figure~\ref{fig: class-wise acc on client0 s}, the updated local model's performance on classes 5 and 8 surpasses that of the initial global model. 
This improvement is due to our proposed loss function, which constrains the local model’s output for vacant classes and suppresses the misclassification of minority samples. 
These adjustments have minimal impact on the learning of majority classes.
Consequently, local models continue to acquire category knowledge from majority classes, such as classes 5 and 8, similar to FedAvg, resulting in enhanced classification accuracy for these classes.

Another interesting observation is that both FedLC~\cite{zhang2022federated} and our method reduce the accuracy of classes 5 and 8 while increasing the accuracy of the remaining classes.
The reason for this behavior is as follows: In FedAvg~\cite{mcmahan2017communication}, the local model often misclassifies vacant and minority classes as majority classes. 
This leads to disproportionately high accuracy for the majority classes and extremely low accuracy for the minority and vacant classes.

FedLC~\cite{zhang2022federated} employs logit weighting to enhance the learning of minority classes, which can result in some majority class samples being misclassified as similar minority classes. 
As a result, this method improves accuracy for minority classes while slightly reducing accuracy for majority classes. 
In contrast, our method introduces vacant-class distillation and logit suppression to substantially mitigate the misclassification of minority and vacant classes as majority classes. 
This approach improves accuracy for vacant and minority classes but may cause some majority class samples to be misclassified as similar vacant or minority classes.
Consequently, while this slightly reduces accuracy for the majority classes, it significantly enhances the overall performance of the local models.

\section{Additional Experimental Results}

\subsection{The Experimental Results on the AG\_news Dataset}
In this subsection, we add the experimental results on the AG\_news dataset with Dirichlet-based ($\beta=0.1$ and $\beta=0.05$) and quantity-based ($s$=2) label skews, as shown in the Tab~\ref{text}, demonstrating our method, FedVLS, consistently outperforms the base-
line methods.
These experiments underscore the versatility
and robustness of FedVLS in real-world federated learning
scenarios facing text classification.

\begin{table}[ht]
\centering
\caption{Performance overview for our method and baselines on the \textbf{AG\_news} dataset with Dirichlet-based ($\beta$=0.05 and $\beta$=0.1) and quantity-based ($s$=2) label skews. \textbf{Bold} is the best result.}
\resizebox{0.4\textwidth}{!}{ 
\begin{tabular}{lccc}
\toprule
Method(venue) & $\beta = 0.1$ & $\beta = 0.05$ & $s = 2$  \\ \cmidrule{1-4}
FedAvg (AISTATS 2017)      & 73.52 & 71.08 & 62.85   \\ 
FedProx (MLSys 2020)      & 75.11 & 71.92 & 64.36   \\ 
FedEXP (ICLR 2023)      & 78.08 & 72.35 & 63.01   \\ 
FedSAM (ICML2022)     & 77.88 & 72.46 & 66.73   \\ 
FedNTD (NeurIPS 2022)      & 79.14 & 75.60 & 69.28   \\ 
FedLMD (MM 2023)      & 82.14 & 77.54 & 71.41   \\ 
FedConcat (AAAI 2024)   & 81.59 & 74.84 & 68.11   \\ 
FedGF (ICML 2024)     & 82.76 & 77.09 & 70.28   \\ 
\textbf{FedVLS (Ours)} & \textbf{87.31} & \textbf{83.19} & \textbf{77.46}  \\
\bottomrule
\end{tabular}
\label{text}
}
\end{table}

\subsection{Compared to Other Knowledge Distillation Methods}
To demonstrate the effectiveness of our vacant-class distillation, we compare it with existing class distillation, normal distillation, DKD~\cite{zhao2022decoupled}, and FedNTD~\cite{lee2022preservation}. 
Similar to FedNTD~\cite{lee2022preservation}, we integrate existing class distillation, normal distillation (KD), and DKD~\cite{zhao2022decoupled} into FedAvg, denoted as FedEKD, FedKD, and FedDKD, respectively.
As shown in Table~\ref{table: distillation}, our method consistently outperforms these approaches.

\begin{table*}[ht]
    \centering
    \caption{The class-wise accuracy for different knowledge distillation methods with Dirichlet-based ($\beta=0.1$) CIFAR10 datasets.}
    \resizebox{0.95\textwidth}{!}{
    \begin{tabular}{lccccccccccc}
    \toprule
         class &  1 &  2 &  3 &  4 &  5 & 6 &  7 &  8 &  9 & 10 & Avg  \\
         \midrule
        global model & 72.20 & 90.90 & 71.30 & 72.10 & 84.40 & 73.60 & 86.20 & 73.80 & 90.60 & 93.80 & 80.89  \\ 
        FedAvg & 0 & 0 & 0 & 44.10 & 0 & 98.40 & 0 & 0 & 97.90 & 95.90 & 33.63  \\ 
        FedEKD & 0 & 0 & 0 & 45.30 & 0 & \textbf{98.80} & 0 & 0 & \textbf{98.90} & 94.90 & 33.79  \\ 
        FedKD & 1.50 & 5.10 & 1.80 & 51.90 & 1.60 & 94.80 & 1.20 & 0 & 97.40 & 95.50 & 35.08  \\
        FedDKD & 3.90 & 34.20 & 16.60 & 52.50 & 9.80 & 94.10 & 14.40 & 0.10 & 97.60 & \textbf{96.00} & 41.92  \\ 
        FedNTD & 8.00 & 38.80 & 22.60 & 58.10 & 11.10 & 96.10 & 12.20 & 0.20 & 98.10 & 94.60 & 43.98  \\ 
        Ours & \textbf{40.40} & \textbf{71.20} & \textbf{39.60} & \textbf{64.60} & \textbf{50.07} & 83.50 & \textbf{54.80} & \textbf{41.30} & 92.30 & 94.32 & 63.21  \\ 
    \bottomrule
    \label{tab: class-wise ACC for distillation}
    \end{tabular}
    }
\end{table*}

\begin{table}
\centering
\caption{Performance overview for different knowledge distillation methods under Dirichlet-based label skews.}
\resizebox{0.48\textwidth}{!}{ 
\begin{tabular}{lcccccc}
\toprule
\multirow{2}{*}{Method} & \multicolumn{2}{c}{\textbf{CIFAR10}} & \multicolumn{2}{c}{\textbf{CIFAR100}} & \multicolumn{2}{c}{\textbf{TinyImageNet}}  \\ \cmidrule{2-7} 
    & $\beta=0.1$ & $\beta=0.05$ & $\beta=0.1$ & $\beta=0.05$ & $\beta=0.1$ & $\beta=0.05$ \\ \midrule
        FedAvg & 82.00 & 62.90 & 66.18 & 62.13 & 39.90 & 35.21  \\ 
        FedEKD & 81.25 & 62.26 & 67.66 & 62.90 & 40.95 & 36.13  \\ 
        FedKD & 82.42 & 64.16 & 67.19 & 63.21 & 41.77 & 36.55  \\ 
        FedDKD & 82.87 & 65.27 & 67.70 & 63.53 & 43.63 & 37.23  \\ 
        FedNTD & 83.23 & 68.71 & 68.00 & 63.71 & 45.11 & 40.65  \\ 
        Ours & \textbf{84.35} & \textbf{75.71} & \textbf{69.02} & \textbf{65.71} & \textbf{47.73} & \textbf{45.23}  \\ 
\bottomrule
\end{tabular}
\label{table: distillation}
}
\end{table}

To investigate the underlying reasons, we further examined the class-wise accuracy of the initial global model and the local models trained using these methods on client 0, whose data distribution is detailed in Table~\ref{tab: data distribution}. 
The specific class-wise accuracy results are presented in Table~\ref{tab: class-wise ACC for distillation}.
FedEKD shows minimal improvement in majority classes but significantly hinders the learning of vacant classes.
FedKD, which uses distillation across all classes, still exhibits low accuracy for vacant classes.
FedDKD adjusts distillation weights for true and not-true classes, while FedNTD applies distillation to not-true classes.
Although these methods improve accuracy for vacant classes, there remains a substantial gap compared to the global model.
Based on these observations, we believe that performing distillation on majority and minority classes will weaken the protection of information for vacant classes.
Therefore, we use vacant-class distillation. 
The results in Table~\ref{tab: class-wise ACC for distillation} further demonstrate that our method significantly enhances the accuracy for vacant classes, finally improving the performance of the local and global models.

\subsection{Combined with Methods for Domain Shift}
Our method is specifically designed to address label skews, making it complementary to approaches that tackle domain skews.
When both domain and label skews are present, our approach can further enhance the performance of methods like FPL~\cite{huang2023rethinking}. 
We have conducted experiments to validate this, with results presented in Table~\ref{domain1} and Table~\ref{domain2}. 
Following the experimental setup in FPL~\cite{huang2023rethinking}, we use the Digits dataset and apply Dirichlet sampling to distribute the data for each domain among six clients.
Under conditions of both domain and label skews, our method significantly improves the performance of PFL~\cite{huang2023rethinking}, demonstrating its effectiveness across different levels of label skews and domain shifts.

\begin{table}[ht]
\centering
\caption{Performance overview for FPL and our method combined with FPL in Dirichlet-based label skews, $\beta$=0.1. \textbf{Bold} is the best result.}
\resizebox{0.48\textwidth}{!}{ 
\begin{tabular}{lccccc}
\toprule
 Method & \textbf{MNIST} & \textbf{USPS} & \textbf{SVHN} & \textbf{SYN} & \textbf{AVG}  \\ \cmidrule{1-6}
        FPL & 97.56 & \textbf{98.73} & 85.06 & 94.23 & 93.89  \\ 
        FPL + Ours & \textbf{98.36} & 98.40 & \textbf{86.66} & \textbf{95.38} & \textbf{94.70} \\
\bottomrule
\end{tabular}
\label{domain1}
}
\end{table}

\begin{table}[ht]
\centering
\caption{Performance overview for FPL and our method combined with FPL in Dirichlet-based label skews, $\beta$=0.05. \textbf{Bold} is the best result.}
\resizebox{0.48\textwidth}{!}{ 
\begin{tabular}{lccccc}
\toprule
 Method & \textbf{MNIST} & \textbf{USPS} & \textbf{SVHN} & \textbf{SYN} & \textbf{AVG}  \\ \cmidrule{1-6}
        FPL      & 96.82 & 96.40 & 77.09 & 89.96 & 90.07  \\ 
        FPL + Ours & \textbf{97.75} & \textbf{97.07} & \textbf{82.05} & \textbf{91.74} & \textbf{92.15} \\
\bottomrule
\end{tabular}
\label{domain2}
}
\end{table}

\subsection{Impact of Communication Rounds}
In real-world scenarios, constraints often limit the number of available communication rounds.
To address this, we evaluate the performance of various methods under different communication round limits using the CIFAR10 dataset with skew parameters $\beta=0.1$ and $\beta=0.05$.
The results, presented in Table~\ref{tab: accuracy diff comm}, show that as the number of communication rounds decreases, the accuracy of most methods drops significantly.
However, our method maintains high accuracy even with fewer communication rounds, demonstrating the robustness and efficiency of FedVLS in environments with restricted communication capabilities.

\begin{table}[ht]
\centering
\caption{Results under varying numbers of communication rounds with Dirichlet-based label skews on CIFAR10 dataset.}
\label{tab: accuracy diff comm}
\resizebox{0.48\textwidth}{!}{ 
\begin{tabular}{lccccccc}
\toprule
\multirow{2}{*}{Method(venue)} & \multicolumn{2}{c}{\textbf{40 comm}} & \multicolumn{2}{c}{\textbf{60 comm}} & \multicolumn{2}{c}{\textbf{80 comm}} \\ \cmidrule(lr){2-3}  \cmidrule(lr){4-5} \cmidrule(lr){6-7} 
& $\beta$=0.1 & $\beta$=0.05 & $\beta$=0.1 & $\beta$=0.05 & $\beta$=0.1 & $\beta$=0.05 \\ \midrule
FedAvg (AISTATS 2017) & 74.62 & 53.44 & 78.59 & 56.71 & 80.72 & 59.10 \\
FedProx (MLSys 2020) & 78.59 & 57.67 & 81.63 & 61.84 & 82.88 & 61.96 \\
MOON (CVPR 2021) & 78.23 & 52.84 & 81.73 & 57.11 & 82.91 & 61.35 \\
FedEXP (ICLR 2023) & 75.90 & 54.14 & 79.69 & 55.98 & 81.51 & 60.01 \\
FedLC (ICML 2022) & 75.74 & 53.06 & 77.22 & 53.77 & 80.22 & 55.75 \\
FedRS (KDD 2021) & 79.10 & 60.99 & 81.13 & 63.16 & 82.94 & 64.28 \\
FedSAM (ICML2022) & 69.02 & 50.05 & 75.42 & 55.85 & 78.38 & 60.79 \\
FedNTD (NeurIPS 2022) & \underline{81.26} & 65.75 & \underline{82.23} & 66.48 & 82.95 & 67.91 \\
FedLMD (MM 2023) & 79.99 & \underline{66.72} & 81.77 & \underline{68.14} & \underline{83.01} & \underline{69.87} \\ \cmidrule{1-7} \rowcolor[gray]{0.9}
\textbf{FedVLS (Ours)} & \textbf{82.54} & \textbf{72.90} & \textbf{83.82} & \textbf{74.34} & \textbf{84.30} & \textbf{75.25} \\
\bottomrule
\end{tabular}
}
\end{table}

\subsection{Impact of Joining Rates, Local Epochs, and Client Numbers}
\begin{figure*}[!ht]
  \centering
  \includegraphics[width=\textwidth]{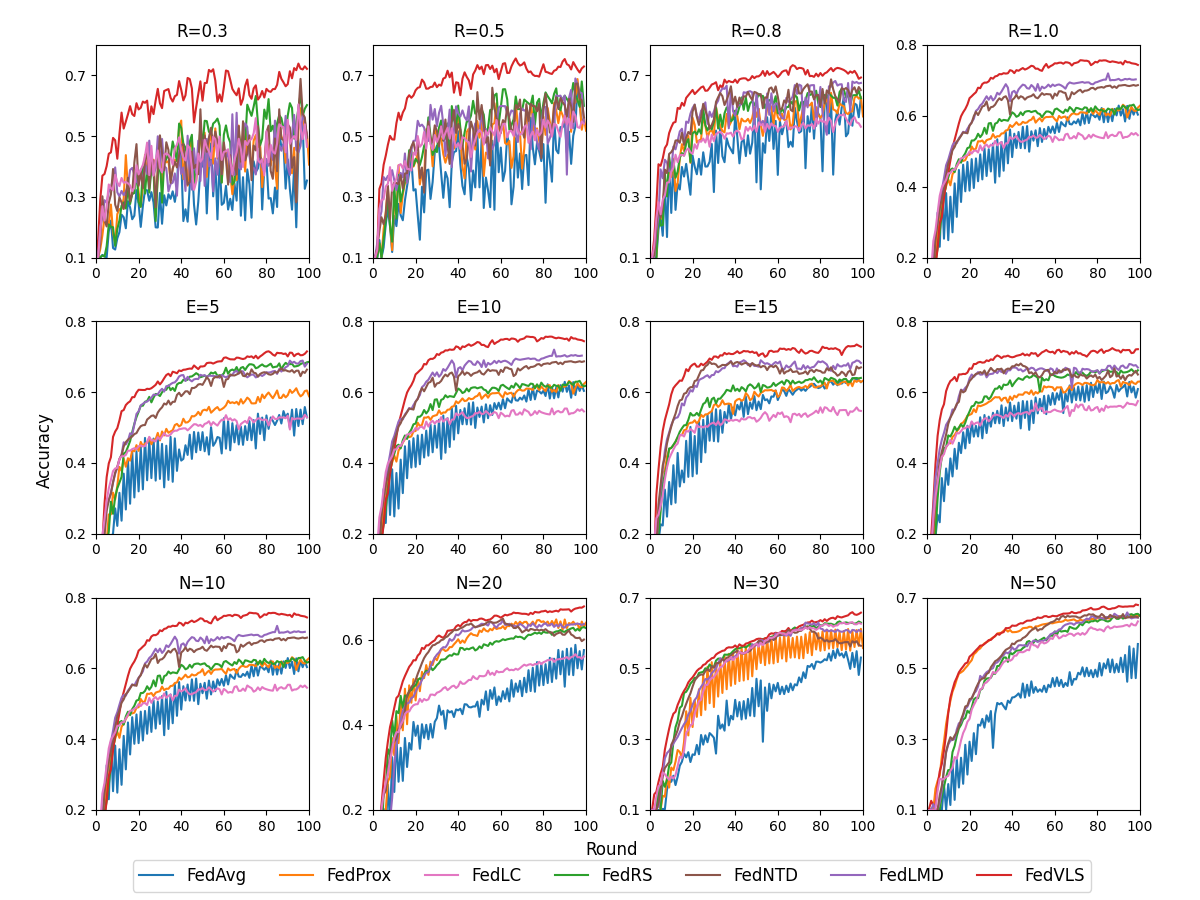}
  \caption{Sensitivity analysis on the client participating rate $\mathbf{R}$, local epochs $\mathbf{E}$, and client numbers $\mathbf{N}$. Each figure separately shows the convergence curve with Dirichlet-based label skews ($\beta=0.05$) on CIFAR10 dataset with $\mathbf{R}$ in $\{0.3, 0.5, 0.8, 1.0\}$, $\mathbf{E}$ in $\{5, 10, 15, 20\}$ and $\mathbf{N}$ in $\{10, 20, 30, 50\}$. }
  \label{fig: ablation_full}
\end{figure*}
Due to space constraints, we included only a portion of the ablation studies on joining rates, local epochs, and client numbers in the main paper. 
Here, we present the complete results.
Specifically, we evaluated joining rates of {0.3, 0.5, 0.8, 1.0}, local epochs of {5, 10, 15, 20}, and client numbers of {10, 20, 30, 50}. 
The experimental results are shown in Figure~\ref{fig: ablation_full}, and the observations are consistent with those presented in the main paper.

As the participation rate decreases, several methods exhibit highly unstable convergence. 
In contrast, our method demonstrates relatively stable convergence, highlighting its robustness to varying participation rates.

Increasing the number of local epochs leads to declining accuracy in the later stages of training for several methods, notably FedNTD~\cite{lee2022preservation}. 
However, our method maintains consistency and improves performance with larger $E$ values, consistently outperforming other methods.

With an increasing number of clients, many methods show slower and less stable convergence.
This is because the larger the number of clients, the greater the damage to model convergence caused by data heterogeneity among clients.
However, our method maintains rapid and stable convergence across varying client numbers, demonstrating the robustness and scalability of our approach.

\subsection{Class-wise Accuracy}
\begin{figure}[ht]
  \centering
  \includegraphics[width=0.48\textwidth]{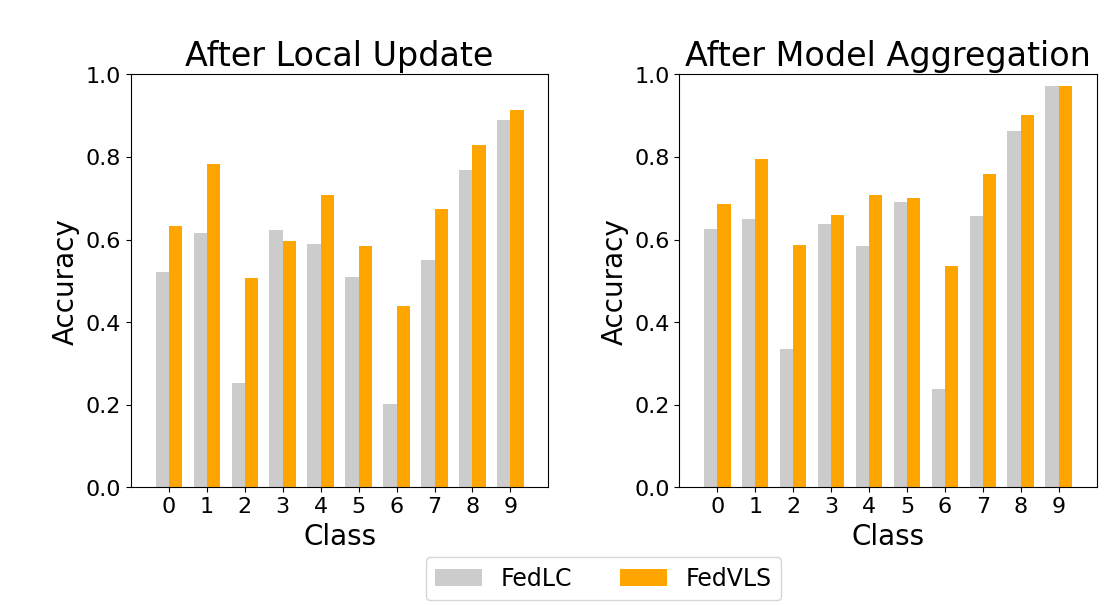}
  \caption{Comparison of class-wise accuracy after local update and after model aggregation with Dirichlet-based label skews ($\beta=0.05$) on CIFAR10 dataset.}
  \label{tab: class-wise-acc}
\end{figure}

To evaluate the effectiveness of our approach, we conduct a comparative analysis of class-wise accuracy before and after local updates using our method, the classic method FedAvg~\cite{mcmahan2017communication}, and the state-of-the-art method FedLC~\cite{zhang2022federated}. 
For a fair comparison, we use the same well-trained federated model as the initial global model, which is then distributed to all clients. 
We train the local models using FedAvg and FedLC, and our method uses the same local data distribution.
As shown in Figure 1 of the main paper, the results align with the observations discussed in the motivation section.
Additionally, we compare the average class-wise accuracy for all clients after local updates and the class-wise accuracy for the aggregated global model of our approach with that of FedLC~\cite{zhang2022federated}, as demonstrated in Figure~\ref{tab: class-wise-acc}.
Our method consistently achieves higher class-wise accuracy compared to FedLC, both after local updates and model aggregation.
These results highlight how our method effectively improves the performance of minority and vacant classes, leading to an overall enhancement in the global model's performance.

\subsection{Model Bias among Clients}
Thanks to the Vacant-class Distillation module, the client model will pay more attention to the vacant classes, which is beneficial to alleviate the model bias among clients. To demonstrate this, we conduct experiments to measure the drift diversity across all client models in the final round following~\cite{li2023effectiveness}. Specially, the drift diversity is defined as follows:
\begin{equation}
{Drift} = \frac {\sum_{i=1}^{N}{\Arrowvert m_i \Arrowvert ^2}}{\Arrowvert \sum_{i=1}^{N}{ m_i }\Arrowvert ^2}, m_i = \omega_i^T - \omega^T
\end{equation}
The results are presented in Table~\ref{tab: drift diversity}. It is evident that our approach effectively mitigates model bias among clients, leading to improved global performance.
\begin{table}[ht]
\centering
\caption{The drift diversity of different method on CIFAR10 datasets with $\beta=0.1$.}
\resizebox{0.48\textwidth}{!}{ 
\begin{tabular}{lcccccc}
\toprule
Method & FedAvg & FedNTD & FedLC &FedVLS (Ours) \\ \midrule
Drift diversity  & 29.73 & 17.85 & 12.11 & \textbf{8.37} \\ 
\bottomrule
\label{tab: drift diversity}
\end{tabular}
}
\end{table}

\section{The Connection between Equation (2) of The Main Paper and FedLC}
Apart from FedLC~\cite{zhang2022federated}, Fedshift~\cite{shen2023federated} also adjusts the logits of model outputs to alleviate model bias caused by imbalanced data distributions. However, they have different forms, so we uniformly represent their loss functions using Eq(2). Nevertheless, during experiments, we train the models according to the original loss function forms as presented in the respective papers. Below, we demonstrate that Eq(2) is positively correlated to the loss function in FedLC~\cite{zhang2022federated}. In Eq(2), 
\begin{align}
\mathcal{L}_{\bf cal} &= -\mathbb{E}_{(x,y) \sim \mathcal{D}_i} \log \left(\frac{p(y) \cdot e^{f(x;{\bf\boldsymbol{\omega}})\left[\it{y}\right]}}{\sum_{c}{p(c) \cdot e^{f(x;{\bf\boldsymbol{\omega}})[\it{c}]}}}\right) \\
 &=  -\mathbb{E}_{(x,y) \sim \mathcal{D}_i} \log \left(\frac{e^{\ln{p(y)}} \cdot e^{f(x;{\bf\boldsymbol{\omega}})\left[\it{y}\right]}}{\sum_{c}{e^{\ln{p(c)}} \cdot e^{f(x;{\bf\boldsymbol{\omega}})[\it{c}]}}}\right) \\
&= -\mathbb{E}_{(x,y) \sim \mathcal{D}_i} \log \left(\frac{e^{\ln{p(y)} + f(x;{\bf\boldsymbol{\omega}})\left[\it{y}\right]}}{\sum_{c}{e^{\ln{p(c)} + f(x;{\bf\boldsymbol{\omega}})[\it{c}]}}}\right),
\end{align}
where $p(y) = \frac{n_y}{n}$, $n_y$ is the number of samples of class $y$ in client $i$, and $n$ is the total number of samples in client $i$. Therefore, Eq(2) can be rewritten in the following form.
\begin{align}
\mathcal{L}_{\bf cal} &= -\mathbb{E}_{(x,y) \sim \mathcal{D}_i} \log \left(\frac{e^{\ln{\left(\frac{n_y}{n}\right)} + f(x;{\bf\boldsymbol{\omega}})\left[\it{y}\right]}}{\sum_{c}{e^{\ln{\left(\frac{n_c}{n}\right)} + f(x;{\bf\boldsymbol{\omega}})[\it{c}]}}}\right) \\
 &= -\mathbb{E}_{(x,y) \sim \mathcal{D}_i} \log \left(\frac{e^{ f(x;{\bf\boldsymbol{\omega}})\left[\it{y}\right] + \ln{n_y} - \ln{n}}}{\sum_{c}{e^{ f(x;{\bf\boldsymbol{\omega}})[\it{c}] + \ln{n_c} + \ln{n}}}}\right)
\end{align}
For different classes within the same client, $n$ remains the same while $n_y$ varies. Therefore, the loss functions for different classes lie in $n_y$ and the output logits. Compared with the loss function in FedLC, 
\begin{equation}
\mathcal{L}_{\bf cal}(y, f(x)) = - \log \left( \frac {e^{f_y(x) - \tau \cdot n_y^{(-1/4)}}}{\sum_{c \neq y}{e^{f_c(x) - \tau \cdot n_y^{(-1/4)}}}}\right) ,
\end{equation}
$\ln{n_y}$ and $- \tau \cdot n_y^{(-1/4)}$ exhibit the same trend as ${n_y}$ changes, therefore they have similar effects on the loss function.

\end{document}